\documentclass[final]{cvpr}

\usepackage{times}
\usepackage{epsfig}
\usepackage{graphicx}
\usepackage{amsmath}
\usepackage{amssymb}

\usepackage[dvipsnames,table]{xcolor}
\usepackage[utf8]{inputenc} %
\usepackage[T1]{fontenc}    %
\usepackage{xurl}
\usepackage{booktabs}       %
\usepackage{amsfonts}       %
\usepackage{nicefrac}       %
\usepackage{microtype}      %
\usepackage{mathtools}  
\usepackage{mathrsfs}
\usepackage{thmtools}
\usepackage{thm-restate}
\usepackage{comment}
\usepackage{parskip}
\usepackage{siunitx}
\usepackage{etoolbox}
\usepackage{enumitem}
\usepackage{subcaption}
\usepackage{xcolor} 
\usepackage{tabularx}
\usepackage[numbers,sort]{natbib}
\usepackage{longtable} 
\usepackage{supertabular}
\usepackage{multirow}
\usepackage{paralist}

\newcommand\blfootnote[1]{%
  \begingroup
  \renewcommand\thefootnote{}\footnote{#1}%
  \addtocounter{footnote}{-1}%
  \endgroup
}

\newcommand{\rotatez}[0]{{\tt\footnotesize RotateZ} }
\newcommand{\flipy}[0]{{\tt\footnotesize FlipY} }

\usepackage[pagebackref=true,breaklinks=true,colorlinks,bookmarks=false]{hyperref}

\def\cvprPaperID{10355} %
\def\confYear{CVPR 2021}

\begin{document}

\title{Pseudo-labeling for Scalable 3D Object Detection}
\author{Benjamin Caine$^{*\dagger}$, Rebecca Roelofs$^{*\dagger}$, Vijay Vasudevan$^{\dagger}$, 
\\Jiquan Ngiam$^{\dagger}$, Yuning Chai$^{\ddagger}$, Zhifeng Chen$^{\dagger}$, Jonathon Shlens$^{\dagger}$\\
$^{\dagger}$Google Brain, $^{\ddagger}$Waymo\\
\texttt{\{bencaine,rofls\}@google.com}\\
}

\maketitle

\begin{abstract}
To safely deploy autonomous vehicles, onboard perception systems must work reliably at high accuracy across a diverse set of environments and geographies. One of the most common techniques to improve the efficacy of such systems in new domains involves collecting large labeled datasets, but such datasets can be extremely costly to obtain, especially if each new deployment geography requires additional data with expensive 3D bounding box annotations. 
We demonstrate that pseudo-labeling for 3D object detection is an effective way to exploit less expensive and more widely available unlabeled data, and can lead to performance gains across various architectures, data augmentation strategies, and sizes of the labeled dataset.
Overall, we show that better teacher models lead to better student models, and that we can distill expensive teachers into efficient, simple students.

Specifically, we demonstrate that pseudo-label-trained student models can outperform supervised models trained on 3-10 times the amount of labeled examples.  Using PointPillars \cite{lang2019pointpillars}, a two-year-old architecture, as our student model, we are able to achieve state of the art accuracy simply by leveraging large quantities of pseudo-labeled data.
Lastly, we show that these student models generalize better than supervised models to a new domain in which we only have unlabeled data, making pseudo-label training an effective form of unsupervised domain adaptation.

\end{abstract}
\section{Introduction}

\begin{figure}[ht!]
  \centering
  \includegraphics[width=0.8\linewidth]{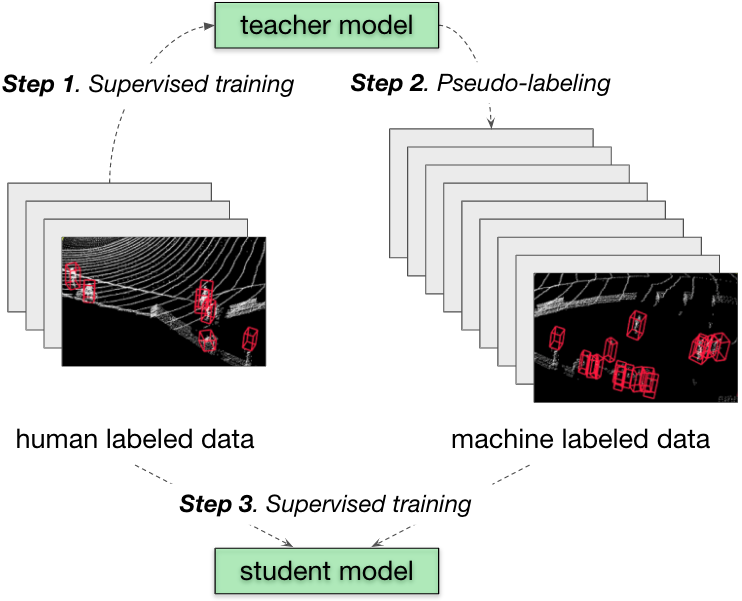}
  \\ \vspace{0.3cm}
  \rowcolors{2}{gray!15}{white}
    \begin{tabular}{ cc |c c c}
    Class & Geography & Baseline & Student & $\Delta$ \\
    \midrule
    Vehicle & SF/MTV/PHX & 49.1 & 58.9  & \textcolor{ForestGreen}{+9.8} \\
    Ped & SF/MTV/PHX & 53.4 & 64.6 & \textcolor{ForestGreen}{+11.2} \\
    \midrule
    Vehicle & Kirkland & 26.1 & 37.2 & \textcolor{ForestGreen}{+11.1} \\
    Ped & Kirkland & 14.5 & 27.1 & \textcolor{ForestGreen}{+12.6} \\
    \end{tabular}
   \caption{\textbf{Pseudo-labeling for 3D object detection.} Top: Training models with pseudo-labeling consists of a three-stage training process. (1) Supervised learning is performed on a {\it teacher} model using a limited corpus of human-labeled data. (2) The teacher model generates pseudo-labels on a larger corpus of unlabeled data. (3) A {\it student} model is trained on a union of labeled and pseudo-labeled data. Bottom: Summary of key results in 3D object detection performance on Waymo Open Dataset \cite{sun2020scalability} with a PointPillars model \cite{lang2019pointpillars}. All numbers report validation set Level 1 difficulty average precision (AP) for vehicles and pedestrians. Both baselines and student models only have access to 10\% of the labeled run segments from original Waymo Open Dataset, which consists of data from San Francisco (SF), Mountain View (MTV), and Phoenix (PHX). We use no labels from the domain adaptation challenge dataset, Kirkland.}
  \label{fig:intro-figure}
  \vspace{-15px}
\end{figure}

Self-driving perception systems typically require sufficient human labels for all objects of interest and subsequently train machine learning systems using supervised learning techniques \cite{thrun2006stanley}. As a result, the autonomous vehicle industry allocates a vast amount of capital to gather large-scale human-labeled datasets in diverse environments \citep{sun2020scalability,geiger2013vision,caesar2020nuscenes}. \iftoggle{cvprfinal}{\blfootnote{$^{*}$ Denotes equal contribution and authors for correspondence.}}{}

However, supervised learning using human-labeled data faces a huge deployment hurdle: while the technique works well on in-domain problems, domain shifts can cause the performance to drop significantly \citep{recht2019imagenet, biggio2017wild,szegedy2013intriguing,hendrycks2018benchmarking}. The reliance of self-driving vehicles on supervised learning implies that the rate at which one can gather human-labeled data in novel geographies and environmental conditions limits wider adoption of the technology. Furthermore, a supervised-learning-based approach is inefficient: for example, it would not leverage human-labeled data from Paris to improve self-driving perception in Rome~\cite{wang2020train}. Unfortunately, we currently have no scalable strategy to address these limitations.

We view the scaling limitations of supervised learning as a fundamental problem, and we identify a new training paradigm for adapting self-driving vehicle perception systems to different geographies and environmental conditions in which human-labeled data is limited or unavailable. We propose leveraging ideas from the literature on semi-supervised learning (SSL), which focuses on the low label regime, and boosts the performance of state-of-the-art models by leveraging unlabeled data. In particular, we employ a pseudo-labeling approach \citep{lee2013pseudo,scudder1965probability, mclachlan1975iterative} to generate labeled data on additional datasets and find that such a strategy leads to significant boosts in performance on 3D object detection (Figure \ref{fig:intro-figure}).

Additionally, we systematically investigate how to structure pseudo-label training to maximize model performance.  We identify nuances not previously well understood in the literature for how best to implement pseudo-labeling and develop simple yet powerful recommendations for how to extract gains from it. Overall, our work demonstrates a viable method for leveraging unsupervised data -- particularly from other domains -- to boost state-of-the-art performance on in-domain and out-of-domain tasks. To summarize our contributions:
\begin{compactitem}
    \item We show pseudo-labeling is extremely effective for 3D object detection, and provide a systematic analysis of how to maximize it's performance benefits.
    \item We demonstrate that pseudo-label training is effective and particularly useful for adapting to new geographical domains for autonomous vehicles.
    \item By optimizing the pseudo-label training pipeline (keeping both the architecture and labeled dataset fixed), we achieve state-of-the-art \textit{test set} performance among comparable models, with 74.0 L1 AP for Vehicles and 69.8 L1 AP for Pedestrians, a gain of 5.4 and 1.9 AP respectively over the same supervised model.  
\end{compactitem}

\section{Related Work}
\textbf{Semi-supervised learning.} Semi-supervised learning (SSL) is an approach to training that typically combines a small amount of human-labeled data with a large amount of unlabeled data
\citep{li2010optimol,papandreou2015weakly,radosavovic2018data,billion_large_scale}.
\textit{Self-training} refers to a style of SSL in which the predictions of a model on unlabeled data, termed \textit{pseudo-labels} \citep{lee2013pseudo}, are used as additional training data to improve performance \citep{scudder1965probability, mclachlan1975iterative}. 
Several variants of self-training exist in the literature. Noisy-Student \citep{xie2020self} uses a smaller, less noised teacher model to generate pseudo-labels, which are used to train a larger, noised student model, and the authors suggest performing multiple iterations of this process. 
FixMatch \citep{sohn2020fixmatch} combines self-training with consistency regularization \citep{sajjadi2016regularization, laine2016temporal}, a technique that applies random perturbations to the input or model to generate more labeled data.
In prior work, self-training has been successfully applied to tasks such as speech recognition \citep{kahn2020self, park2020improved}, image segmentation \citep{Chen2020Leveraging}, and 2D object detection in camera imagery \citep{rosenberg2005semi, zoph2020rethinking, sohn2020simple} and video sequences \citep{chen2020semi}.

\textbf{3D object detection.}
Though several architectural innovations have been proposed for 3D object detection \citep{yang2018pixor,zhou2018voxelnet,yan2018second,luo2018fast,yang2018hdnet,ngiam2019starnet}, a recent focus has been on techniques that improve \textit{data efficiency}, or the amount of data required to reach a certain performance. 
Data augmentation designed for 3D point clouds can significantly boost performance (see references in \citep{li2020pointaugment,cheng2020improving}), and techniques to automatically learn appropriate data augmentation strategies have been shown to be $10$ times more data efficient than baseline 3D detection models \cite{cheng2020improving,li2020pointaugment}. Concurrent to our work, \cite{wang2020multi} shows gains applying knowledge distillation \cite{hinton2015distilling} to 3D detection, distilling a multi-frame model's features to a single-frame model in \textit{feature space}, whereas we apply knowledge distillation in \textit{label space}.

Several recent works also propose improving data efficiency by using \textit{weak supervision} to augment existing labeled data: \citep{tang2019transferable} incorporates existing 3D box priors to augment 2D bounding boxes, and \cite{meng2020weakly} similarly generates additional 3D annotations by learning appropriate augmentations for labeled object centers. Finally, an automatic 3D bounding box labeling process is proposed by \cite{yang2021auto4d}, which uses the full object trajectory to produce accurate bounding box predictions, though they don't show training results with these auto labels.

We view many of the techniques to improve data efficiency as complementary to our work, as improvements in either model architectures or data efficiency will provide additive performance benefits.

\textbf{SSL for 3D object detection.} Two prior works apply Mean Teacher \cite{tarvainen2018mean} based semi-supervised learning techniques to 3D object detection \cite{wang20203dioumatch, zhao2020sess} on the indoor RGB-D datasets ScanNet \cite{dai2017scannet} and SUN RGB-D \cite{song2015sun}. SESS \cite{zhao2020sess} trains a student model with several consistency losses between the student and the EMA-based teacher model, while 3DIoUMatch \cite{wang20203dioumatch} proposes training directly on the pseudo labels after filtering them via an IoU prediction mechanism. In contrast, we forgo a Mean-Teacher-based framework, finding separate teacher and student models to be practically advantageous, and we showcase performance on 3D LiDAR datasets designed to train self-driving car perception systems.

\textbf{Domain adaptation.} Robustness to geographies and environmental conditions is critical to making self-driving technology viable in the real world \cite{thrun2006stanley}. Recently, one group studied the task of adapting a 3D object detection architecture across self-driving vehicle datasets (e.g. \cite{geiger2013vision,houston2020one,sun2020scalability}), and reported significant drops in accuracy when training on one dataset and testing on another \cite{wang2020train}. Interestingly, such drops in accuracy could be attributed to differences in car sizes and are partially reduced by accounting for these size differences.
In parallel, other recent work reports notable drops in accuracy across geographies within a single dataset \cite{sun2020scalability} (see Table 9). However, unlike the former work, those drops in accuracy in this latter work cannot be accounted for by differences in car sizes \footnote{We found the average width and length of vehicles in Kirkland and the Waymo Open Dataset to be quite similar. For instance, in the validation splits of the Waymo OD and Kirkland datasets, we measured similar average lengths (4.8m vs 4.6m) and average widths (2.1m vs 2.1m) across O($10^4$) objects. These discrepancies are markedly less than those described in \cite{wang2020train}.}. In our work, we experiment on this single dataset and are able to mitigate drops in accuracy across geographies.

We focus on one of the open challenges for the Waymo Open Dataset \footnote{\,\url{https://waymo.com/open/challenges}}: accurate 3D detection in a new city (Kirkland) with changing environmental conditions (rain) and limited human-labeled data. 
Currently, the state-of-the-art architecture for the Kirkland domain adaptation task \citep{ding20201st} employs a single-stage, anchor-free and NMS-free 3D point cloud object detector equipped with multiple enhancements including features from 2D camera neural networks, powerful data augmentation, frame stacking, test time ensembling, and point cloud densification (but no pseudo-labeling).
We do not implement these full set of enhancements, yet our baseline implementation achieves similar performance to their baseline architecture \cite{ge2020afdet}, instead focusing on accuracy and robustness gains that can be achieved by leveraging a large amount of unlabeled data. %

\section{Methods}

\begin{figure}[t!]
  \centering
  \includegraphics[width=1.0\linewidth]{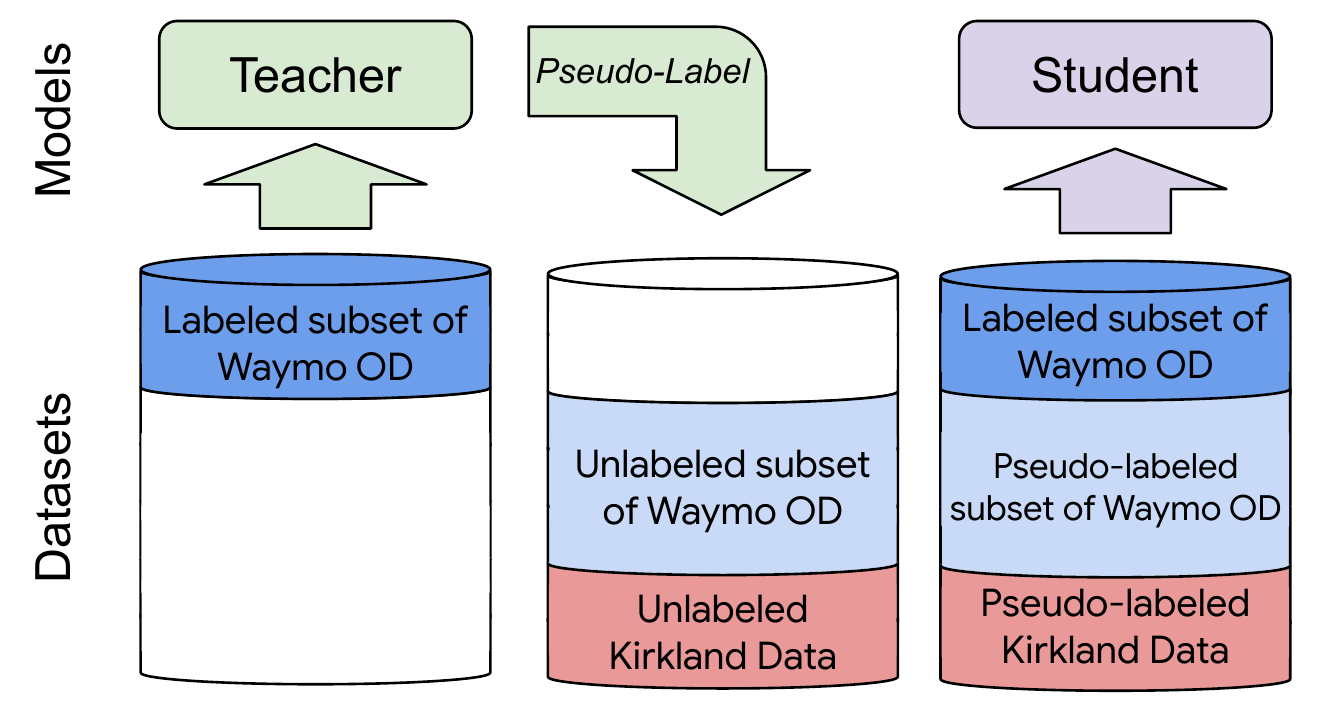}
  \caption{\textbf{Experimental setup.} We conduct our experiments on the Waymo Open Dataset \cite{sun2020scalability}, where we artificially divide the dataset into labeled and unlabeled splits. We always treat run segments from Kirkland as unlabeled (even though a subset are labeled) and select subsets (e.g. 10\%, 20\%, ...) of the original Waymo Open Dataset run segments to train the teacher. We use the teacher to pseudo-label all unseen run segments, and then train a student on the union of labeled and pseudo-labeled run segments. Finally, we evaluate both teacher and student models on the original Waymo Open Dataset and Kirkland validation splits.}
  \label{fig:overview}
  \vspace{-10px}
\end{figure}

Our pseudo-labeling process (Figure \ref{fig:overview}) consists of three stages: training a teacher on labeled data, pseudo-labeling unlabeled data with said teacher, and training a student on the combination of the labeled and pseudo-labeled data. We perform and evaluate all of our experiments on the Waymo Open Dataset (version 1.1) \cite{sun2020scalability} and the domain adaptation extension. We implement students and teachers as PointPillars \cite{lang2019pointpillars} models using open-source implementations \footnote{\,\url{https://github.com/tensorflow/lingvo/}}, which are the baselines used by \cite{ngiam2019starnet,sun2020scalability,cheng2020improving, han2020streaming}. 

\subsection{Data Setup}
The Waymo Open Dataset \cite{sun2020scalability} is organized as a collection of \textbf{run segments}. Each run segment is a $\sim$200 frame sequence of LiDAR and camera data collected at 10Hz. These run segments come from two sets: the original Waymo Open Dataset, which has 798 labeled training run segments collected in San Francisco, Phoenix, and Mountain View, and the domain adaptation benchmark, which has 80 labeled and 480 unlabeled training run segments from Kirkland. Both datasets contain 3D bounding boxes for {\it Pedestrian}, {\it Vehicle}, and {\it Cyclist}, but, due to the low number of Cyclists in the data, we focus on the Pedestrian and Vehicle classes.  

In our experiments, we treat all the Kirkland run segments as unlabeled data (even though labels do exist for 80 run segments). Our setup is similar to unsupervised domain adaptation, where only unlabeled data is available in the ``target'' domain, giving us a measure of how well the gains in accuracy on the Waymo Open Dataset generalize to a new domain \footnote{In addition to geographical nuances, Kirkland has notably different weather conditions, e.g. clouds and rain, than San Francisco, Phoenix, and Mountain View.}.  In addition, our setup emulates a common scenario in which a practitioner has access to a large collection of unlabeled run segments and a much smaller subset of labeled run segments. 

In order to study the effect of labeled dataset size, we randomly sample smaller training datasets from the Waymo Open Dataset. Because run segments are typically labeled efficiently as a sequence, we treat each run segment as either comprehensively labeled or unlabeled, and we sample based on the \textit{run segment IDs}, instead of individual frames. For example, selecting 10\% of the original Waymo Open Dataset corresponds to selecting 10\% of the run segments, i.e. 79 run segments, which provides $\sim$15,700 frames. If we were to instead randomly select 10\% of \textit{frames}, we would make the task artificially easier, as neighboring frames would be highly correlated, especially if the autonomous vehicle is moving slowly.

\subsection{Model Setup}
All experiments use PointPillars \cite{lang2019pointpillars} as a baseline architecture due to its simplicity, accuracy, and inference speed (see Appendix \ref{sec:pillar_details} for architecture details). To explore the impact of teacher accuracy, we use wider and multi-frame PointPillars models as teachers. To make the models wider, we multiply all channel dimensions by either $2\times$ or $4\times$. To make a multi-frame teacher, we concatenate the point clouds from each frame with its previous $N-1$ frames transformed into the last frame's coordinate system.

\subsection{Training Setup}

Our training setup mirrors \cite{sun2020scalability, cheng2020improving}. We use the Adam optimizer \cite{kingma2014adam} and train with an exponential decay schedule on the learning rate. All teachers and students are trained with the same schedule, but the length of an epoch for teacher and student models differ because the teacher is trained on less data than the student. 

We use data augmentation strategies such as world rotation and scene mirroring, which showed strong improvement over not using augmentations.  Table  \ref{tab:teacher_augmentations} provides an ablation study for these augmentations. Unless otherwise stated, all other training hyperparameters remain fixed between teacher and student.  See Appendix \ref{sec:training_details} for additional details on the training setup.

\subsection{Pseudo-Label Training}
Pseudo-label training begins by training a teacher model using standard supervised learning on a labeled subset of run segments.  
Once we train the teacher, we select the best teacher model based on validation set performance on the Waymo Open Dataset and use to pseudo-label the unlabeled run segments.
Next, we train a student model on the \textit{same} labeled data the teacher saw, plus all the pseudo-labeled run segments. The mixing ratio of labeled to pseudo-labeled data is determined by the percentage of data the teacher was trained on. 

We filter the pseudo-labeled boxes to include only those with a classification score exceeding a threshold, which we select using accuracy on a validation set. We find a classification score threshold of 0.5 works well for most models, but a small subset of models (generally multi-frame Pedestrian models, which are poorly calibrated and systematically under-confident) benefit from a lower threshold.
Finally, we evaluate the student's performance on the Waymo Open Dataset and Kirkland validation sets, where we always report Level 1 (L1) average precision (AP).
\section{Results}
\label{sec:results}
Using the Vehicle class, we first explore the relationship between teacher and student performance on the Waymo Open Dataset for various teacher configurations, and then evaluate generalization to Kirkland.
Next, for both Vehicles and Pedestrians, we distill increasingly larger teachers into small, efficient student models, yielding large gains in accuracy with no additional labeled data or inference cost. Finally, we scale up these experiments with two orders of magnitude more unlabeled data, further demonstrating the efficacy of pseudo-labeling. We also describe some negative results where we discuss some ideas we thought should work, but did not.

\subsection{Better teachers lead to better students.} \label{better_teachers_better_students}
\begin{figure}[t!]
  \centering
  \includegraphics[width=0.7\linewidth]{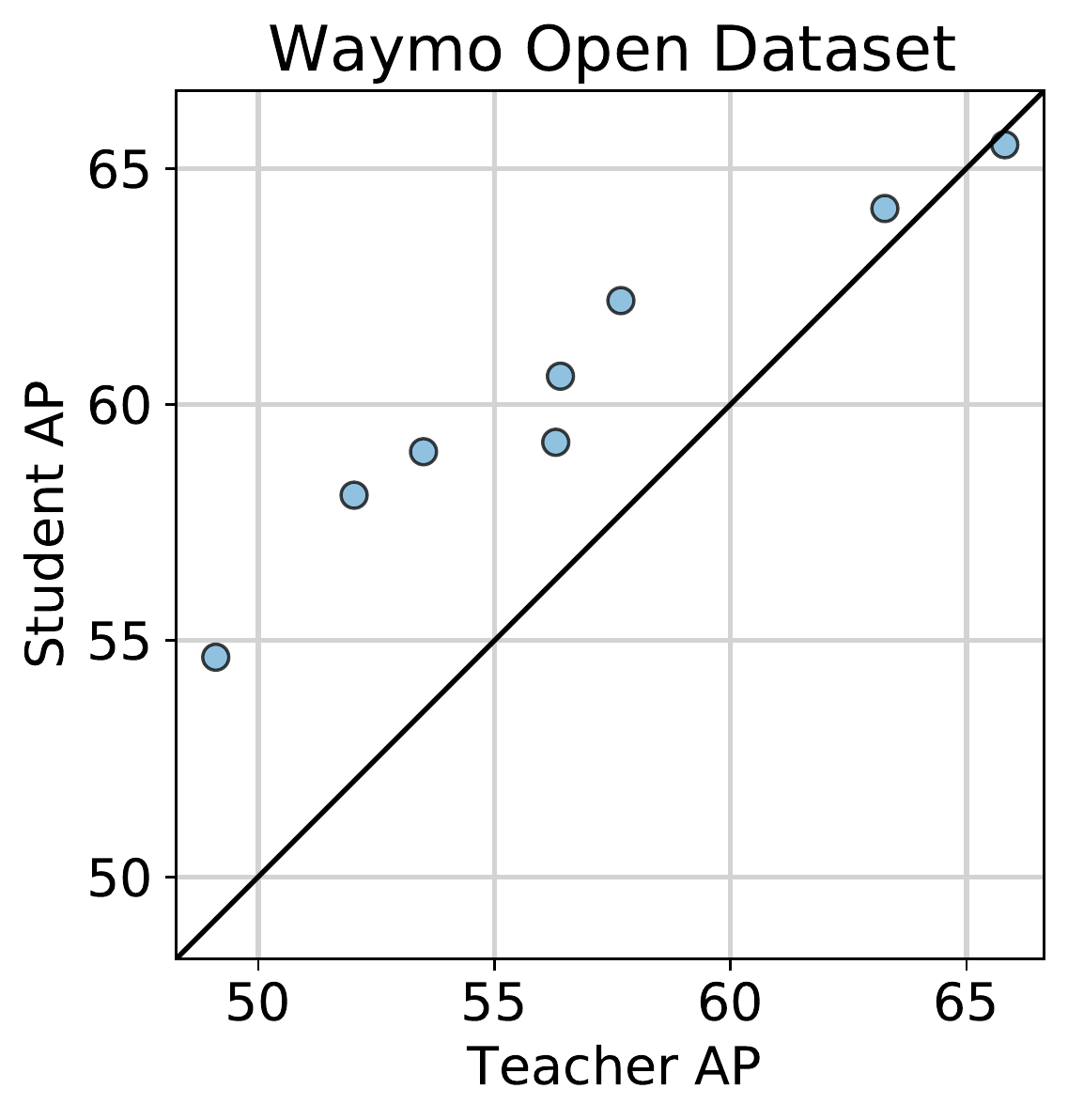}
  \vspace{-5pt}
  \caption{
    \textbf{Better teachers lead to better students.}
    We plot Level 1 AP on the Waymo Open Dataset \textit{validation set} for Vehicles.
    When controlling for labeled dataset size, architecture, and training setup between teachers and students, teachers with a higher AP generally produce students with a higher AP.
    }
  \label{fig:student_vs_teacher}
  \vspace{-12px}
\end{figure}
To understand how teacher performance impacts student performance, we control the accuracy of the teacher by varying the amount of labeled data, the teacher's width, and the strength of teacher training data augmentations. All experiments in this section are evaluated on Vehicles. In Figure \ref{fig:student_vs_teacher}, we show student-versus-teacher performance for teacher and student models with the same amount of \textit{labeled} data, equivalent architectures, and equivalent training setups.

In general, higher accuracy teachers produce higher accuracy students. A relevant question is then, what techniques are most effective for improving teacher accuracy? To answer this, we evaluate each modification in turn on the Waymo Open Dataset. Appendix \ref{apx:kirkland_teacher_configs} shows the corresponding experiments when evaluating on the Kirkland dataset.

\begin{figure}[hb!]
  \newcommand{\ppwm}{0.48}
  \centering
    \includegraphics[width=0.95\linewidth]{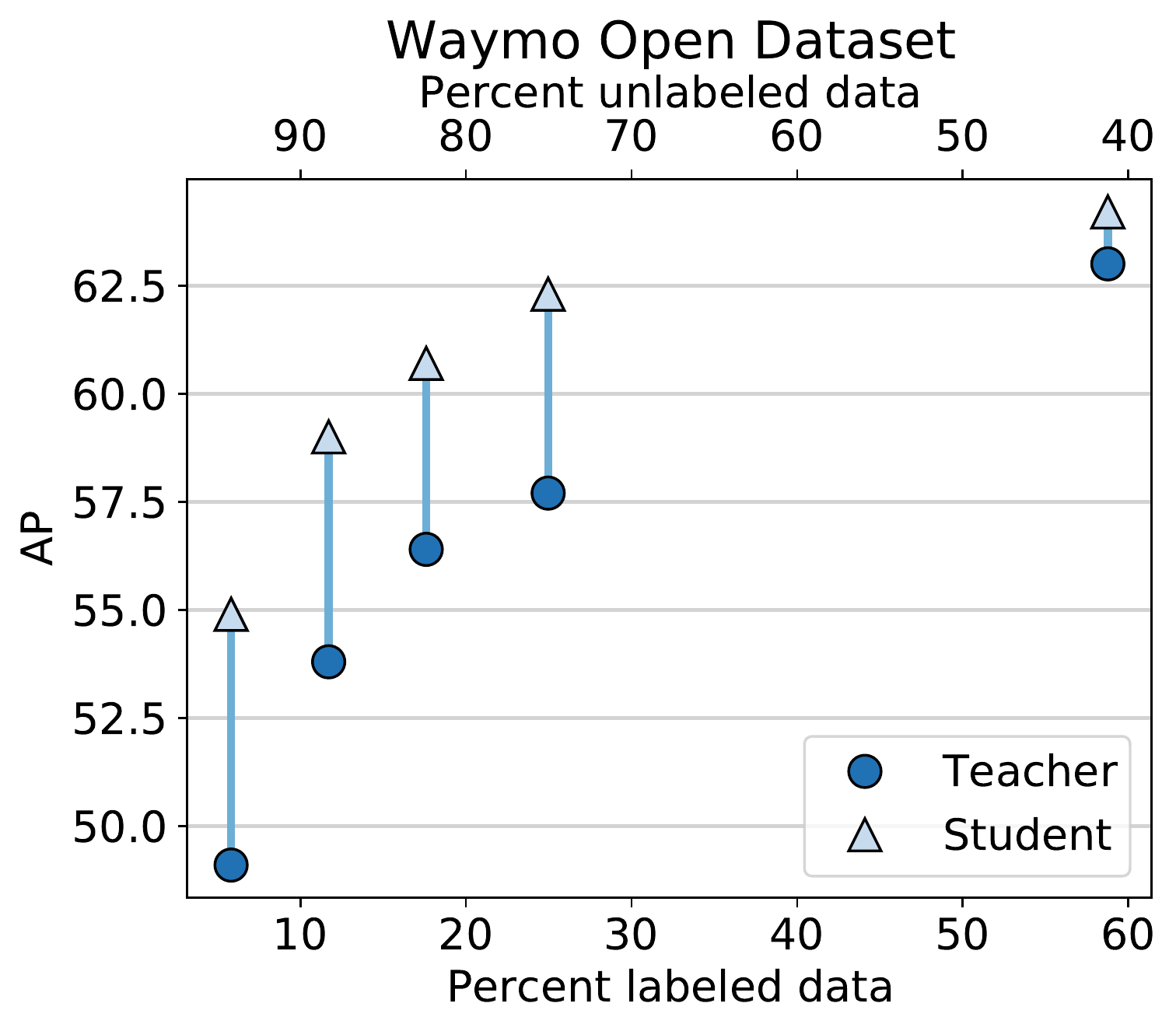}
  \vspace{-5pt}
  \caption{
    \textbf{Pseudo-label training is most effective when the ratio of labeled to unlabeled data is small.} Teacher and student L1 AP on the Waymo Open Dataset \textit{validation set} for the Vehicle class versus overall percent labeled data.
    }
  \label{fig:teacher_overall_percent_labeled_data}
  \vspace{-6pt}
\end{figure}

\textbf{Amount of labeled data.} Compared to adding data augmentations or increasing teacher width, increasing the amount of labeled data yields the largest improvements. Figure \ref{fig:teacher_overall_percent_labeled_data} shows that increasing the fraction of labeled data improves both teacher and student performance, but the student gains diminish as the amount of unlabeled data decreases.

Note that Figure \ref{fig:teacher_overall_percent_labeled_data} shows the \textit{overall} percent labeled (bottom axis) and unlabeled data (top axis) when we combine the 798 Waymo Open Dataset and 560 Kirkland run segments.  Using 100\% of the labeled data from the Waymo Open Dataset corresponds to having roughly 59\% of the overall data labeled, and we give teachers access to 10\%, 20\%, 30\%, 50\% of 100\% of the labels in the Waymo Open Dataset in this experiment, allowing us to evaluate the effect of having access to x\% human labeled and (1-x)\% pseudo-labeled data from the Waymo Open Dataset.

\begin{table}[t!]
    \centering
    \rowcolors{2}{gray!15}{white}
    \begin{tabular}{ c |c  c c }
      & \multicolumn{3}{c}{\textbf{Waymo Open Dataset L1 AP}} \\ %
    \midrule
    Teacher Augmentation & Teacher & Student & $\Delta$ \\
    \midrule
    None    & 56.3   & 62.2   & \textcolor{ForestGreen}{+5.9}  \\
    \flipy   & 60.1  & 63.6  & \textcolor{ForestGreen}{+3.5}  \\
    \rotatez & 61.4  & 63.3  & \textcolor{ForestGreen}{+2.1}  \\
    \rotatez + \flipy & 63.0  & 64.2   & \textcolor{ForestGreen}{+1.2} \\
    \end{tabular}
    \caption{\textbf{Stronger teacher augmentations lead to additive gains in student performance.} We increasing the strength of teacher augmentations for a 1$\times$ width teacher model trained on 100\% of the Waymo Open Dataset, while fixing the student to be a 1$\times$ width model trained with both \rotatez and \flipy augmentations. We report L1 \textit{validation set} Vehicle AP on the Waymo Open Dataset. $\Delta = \text{Student AP} - \text{Teacher AP}$.}
    \label{tab:teacher_augmentations}
    \vspace{-10pt}
\end{table}

\textbf{Data augmentation.}
We find that adding data augmentation does lead to modest additional gains, mirroring the observations in \cite{zoph2020rethinking}, as long as it is applied to both the teacher and the student. 
In Table \ref{tab:teacher_augmentations}, we show that one way to generate stronger teacher models (and thus better students) is through stronger data augmentations. 

Although \cite{xie2020self} emphasizes the importance of noising the student model, we found empirically that pseudo-label training can show gains even \textit{without} data augmentation (see Appendix \ref{apx:data_augmentation_necessary} for full results).

\begin{figure}[t!]
  \newcommand{\ppwm}{0.48}
  \centering
  \includegraphics[width=0.95\linewidth]{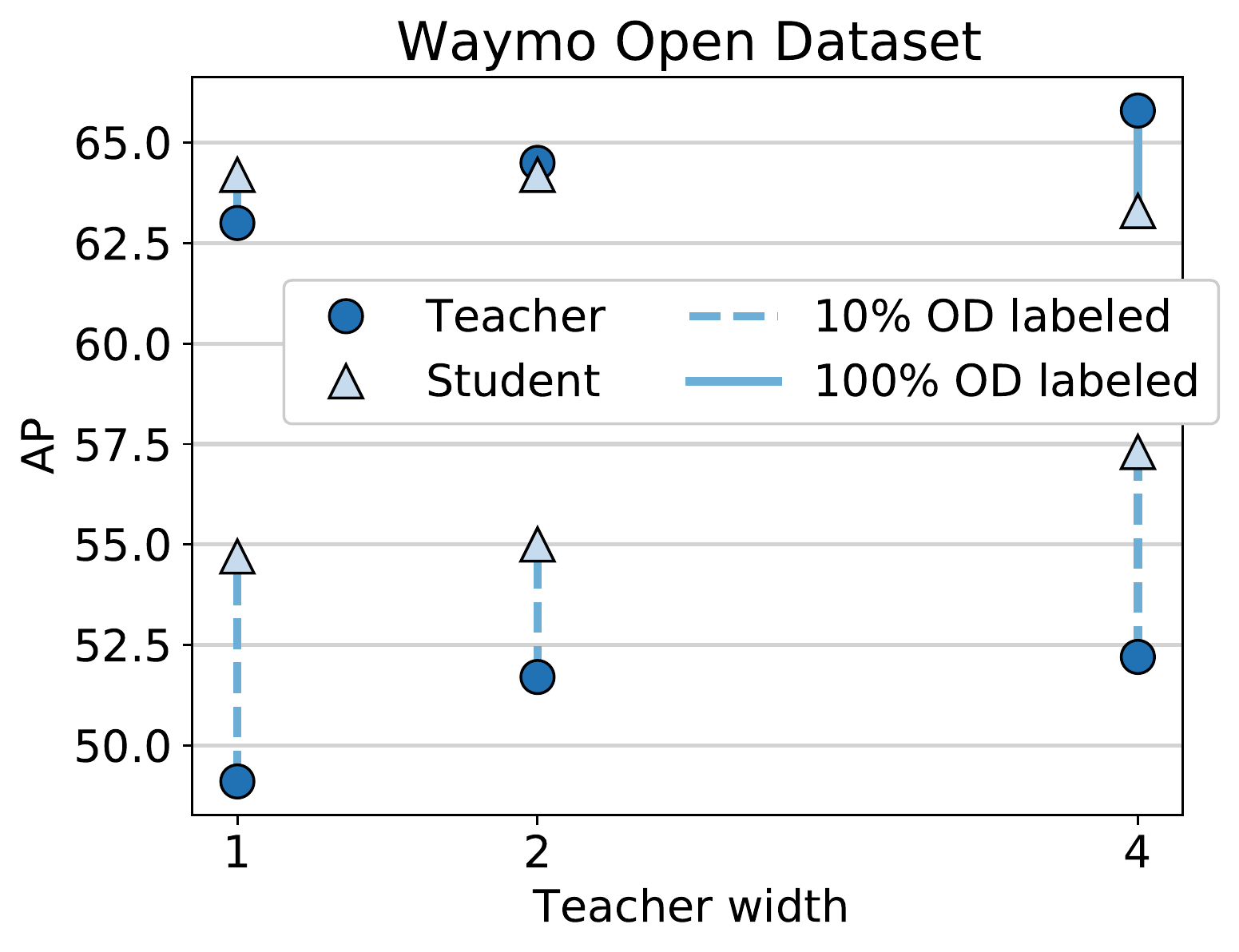}
  \vspace{-10pt}
  \caption{
    \textbf{Increasing teacher width leads to better students when labeled data is limited.}
    We increase teacher width while fixing the student width at 1$\times$ and compare L1 AP for Vehicle models on the Waymo Open Dataset \textit{validation set}. The teacher is trained on labeled data from either 10\% or 100\% of the original Waymo Open Dataset (bottom and top points, respectively).
    When the ratio of labeled to unlabeled data is small, student accuracy improves as the teacher gets wider. However, this effect disappears when the amount of pseudo-labeled data is small. 
    }
  \vspace{-10pt}
  \label{fig:teacher_width}
\end{figure}

\textbf{Teacher width.} 
An additional way to generate better teachers is through scaling the model size (parameter count). Because the teacher and student are different models, they can be of different sizes, architectures, or configurations.
One useful strategy involves distilling a large, expensive offline model's performance into a small, efficient production model. 
In Figure \ref{fig:teacher_width}, we vary the teacher width (1$\times$, 2$\times$, or 4$\times$) by multiplying all its channel dimensions while keeping the student width fixed at 1$\times$.  
We evaluate performance under two different fractions of available labeled data (10\% or 100\% of the original Waymo Open Dataset).

When only 10\% of the original Waymo Open Dataset is labeled, \emph{the 1$\times$ width students outperform their wider teachers}, in contrast to the findings in \cite{xie2020self} that require the student to be equal or larger than the teacher, suggesting that in the low labeled data regime, this may not be as important. However, when 100\% of the original Waymo Open Dataset is labeled, the 1$\times$ width student can no longer outperform the 4$\times$ width teacher on the original Waymo Open Dataset.

\textbf{Ratio of labeled to unlabeled data.}
In our results, we found that in the setting where the ratio of labeled to unlabeled data is high -- using 100\% of the original Waymo Open Dataset's labels (798 segments) and only pseudo-labeling Kirkland's data (560 segments) -- the student gains compared to the teacher diminish, and the student is unable to outperform a wider teacher.

One hypothesis is that the lack of improvement is due to the small amount of unlabeled data that the student can benefit from.  We test this hypothesis by using two orders of magnitude more unlabeled data in Section \ref{large_scale_unlabeled_data}.

\subsection{Generalization to Kirkland}
\begin{figure}[h]
  \newcommand{\ppwm}{0.47}
  \centering
  \begin{subfigure}{\ppwm\textwidth}
    \includegraphics[width=0.95\linewidth]{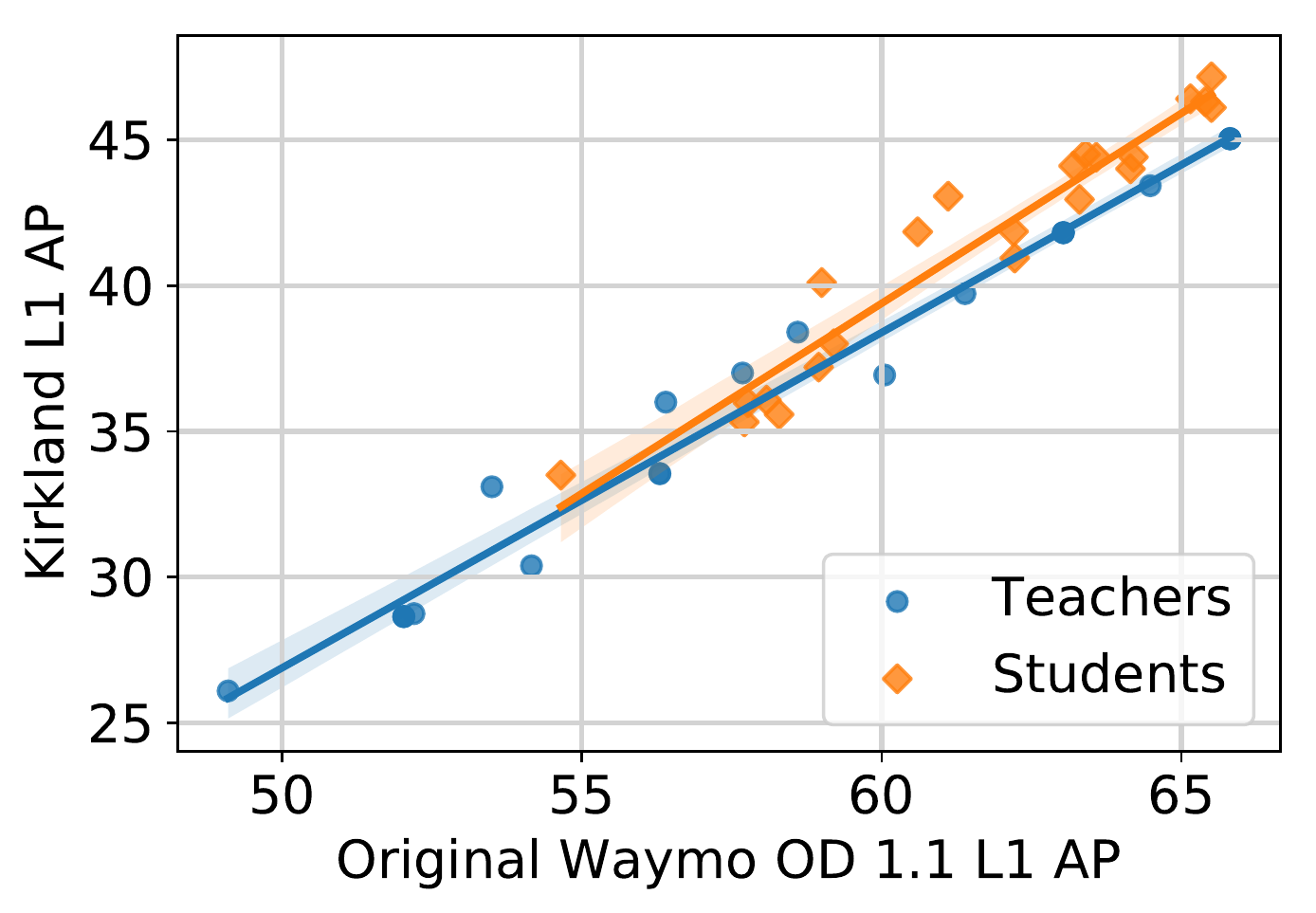}
  \end{subfigure}
    \vspace{-8pt}
  \caption{
    \textbf{Pseudo-labeling improves performance on unlabeled geographic domain.} Stronger models on the original Waymo Open Dataset are also better on the Kirkland dataset (where we only have unlabeled data). Moreover, student models trained with pseudo-labeling generalize better to Kirkland than normally supervised teacher models.}
  \label{fig:od_vs_kirkland}
  \vspace{-7px}
\end{figure}
In order to measure generalization to new geographies and environmental conditions, we evaluate all models on the Kirkland domain adaptation challenge dataset.  
The weather in Kirkland is rainier than the weather in the cities that comprise the Waymo Open Dataset, which increases the level of noise in the LiDAR data. 
We plot the model's performance on the Kirkland dataset versus the model's performance on the Waymo Open Dataset for both teacher and student models in Figure \ref{fig:od_vs_kirkland}. We observe a clear linear relationship between the model's performance on the Waymo Open Dataset and the model's performance on Kirkland, implying that a model’s accuracy on the Waymo Open Dataset can almost perfectly predict accuracy on the Kirkland dataset. Overall, the Kirkland performance is much lower than the Waymo Open Dataset performance, which we suspect is due to an underlying data distribution difference and the fact that we only use labeled data from the Waymo Open Dataset in training.

Interestingly, the slope of the linear relationship changes depending on whether the model is a teacher or student; the student models have a slightly higher slope than the teacher models, indicating that the student models are generalizing better to the Kirkland dataset.  We find that the difference in slope is statistically significant by using an Analysis of Covariance (ANCOVA) test, which evaluates whether the means of our dependent variable (Kirkland AP) are equal across our categorical independent variable (whether the model is a student or not), while statistically controlling for accuracy on the Waymo Open Dataset. We find an F-score of 12.9, giving us a p-value less than 0.001, which is lower than 0.05 (the significance level for 95\% confidence), leading us to reject the null hypothesis.  Since the student models have a slightly higher Kirkland AP for a given Waymo Open Dataset AP, we conclude that the student models are slightly more robust to the Kirkland distribution shift.

\subsection{Pushing labeled data efficiency}
\begin{figure*}[h]
  \centering
  \includegraphics[width=0.9\linewidth]{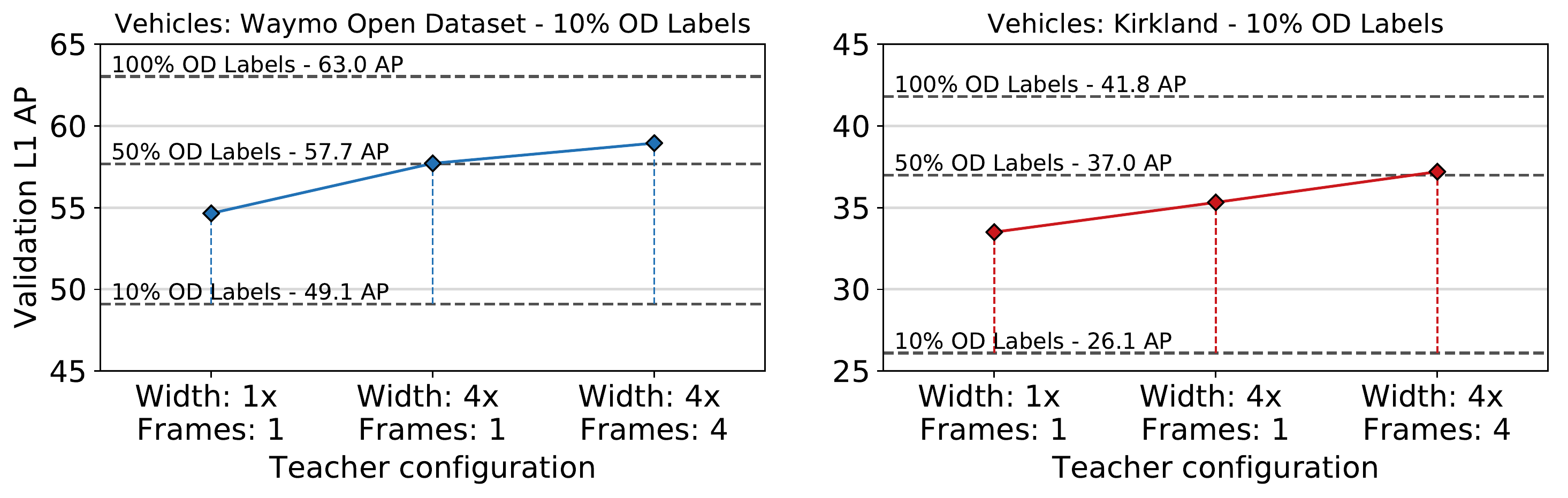}
  \includegraphics[width=0.9\linewidth]{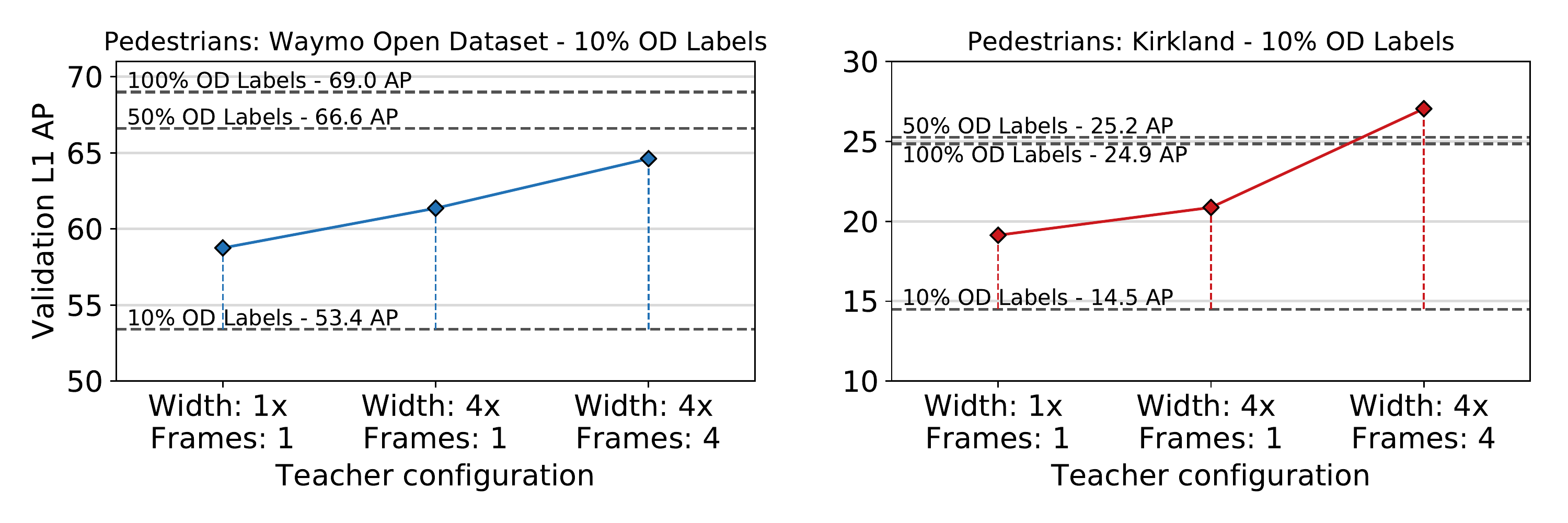}
  \vspace{-0.4cm}
  \caption{\textbf{Increasingly large teachers distill into small, accurate students.} Increasing the width or number of frames for the teacher impacts performance on a fixed size student (1$\times$ width, 1 frame) in the low label  (10\% of run segments) regime for vehicles (top) and pedestrians (bottom). Training a large, expensive teacher model, and distilling the teacher into a small, efficient student is an efficient tactic. We present results for Waymo Open Dataset (left) and Kirkland (right).}
  \label{fig:finale_10pct}
  \end{figure*}

For practitioners, an important question is \textit{"How do I make the most accurate model given a fixed inference time budget and fixed amount of labeled data?"}  We assume that autonomous vehicle practitioners have more unlabeled than labeled data due to the relative ease of collecting vs. comprehensively labeling data. In our experiment, we show that better teachers still lead to better students, even as we make larger, more accurate teacher models, and that distilling an expensive, impractical offline model into an efficient, practical production model via pseudo-labeling is an effective technique.
Additionally, we show via strong Kirkland validation set results (a domain where we use no labeled data) that pseudo-labeling is an effective form of unsupervised domain adaptation. 

We improve the teacher by both scaling its width to 4$\times$ and concatenating up to four LiDAR frames as input. As with all of our experiments, our training setup for students mirrors the teacher except that the student models are always 1$\times$ width, 1 frame. Our results are shown in Figure \ref{fig:finale_10pct} and summarized in Figure \ref{fig:intro-figure}.

For Vehicle models, we find that distilling a 4$\times$ width, 4 frame teacher model into a 1$\times$ width, 1 frame student model, using only 10\% of the original Waymo Open Dataset labels, can match or exceed the performance of an equivalent supervised model trained with 5$\times$ that amount of labeled data. Our Pedestrian model is even more remarkable: using only 10\% of the original Waymo Open Dataset run segment labels, our student model outperforms an equivalent supervised baseline on Kirkland trained on 10$\times$ the amount of labels from the original Waymo Open Dataset. \footnote{Note that we find that the 4$\times$ width, 4 frame Pedestrian models were systematically under-confident, and lowering the pseudo-label score threshold from 0.5 to 0.3 improved results.}

Our results show that unlabeled data in-domain can be vastly more effective than labeled data from a different domain.
Additionally, in Appendix \ref{apx:pushing_labeled_data_efficiency} we show that these results hold when doubling the amount of labeled data.

\begin{table}[h!]
\small
    \centering
    \rowcolors{2}{gray!15}{white}
    \begin{tabular}{ c | >{\centering}m{1.1cm} >{\centering}m{1.2cm} | >{\centering}m{0.4cm} >{\centering}m{0.4cm} >{\centering}m{0.4cm} c}
    Training & OD / Kir & OD / Kir & \multicolumn{4}{c}{OD L1 AP} \\
    Method & \# label & \# pseudo & Veh  & $\Delta$ & Ped & $\Delta$ \\
    \midrule
    baseline & 800 / 0 & 0 / 0 & 63.0 & -- & 69.0 & -- \\
    semi-super & 800 / 0 & \,\,\,\,\,\,0 / 560 & 64.2 & {\footnotesize \textcolor{ForestGreen}{+1.2}} & 69.8 & {\footnotesize \textcolor{ForestGreen}{+0.8}} \\
    semi-super & 800 / 0 & \,\,\,\,0 / 8k & \textbf{65.1} & {\footnotesize \textcolor{ForestGreen}{+2.1}} & 68.8 & {\footnotesize \textcolor{Red}{-0.9}} \\
    semi-super & 800 / 0 & 67k / 8k & \textbf{68.8} & {\footnotesize \textcolor{ForestGreen}{+5.8}} & 70.5 & {\footnotesize \textcolor{ForestGreen}{+1.5}} \\
    \end{tabular}
    \caption{\textbf{Pseudo-labeling increases accuracy in domain.} The number of labels are reported in {\it run segments}. All performance numbers report validation set L1 difficulty AP for the original Waymo Open Dataset with the {\it same 1$\times$ width 1 frame} network architecture. Only the training method varies across each experiment. $\Delta$ indicates the difference in AP with respect to the {\it Baseline} model, which is trained only on the Waymo Open Dataset (OD). {\it Semi-supervised} uses a 4$\times$ width, 4 frame teacher model trained on OD labeled data to provide pseudo-labels and then trains the student on the joint labeled and pseudo-labeled data. We include Kirkland data to show that out of domain data also provides gains, but not as large.}
    \label{tab:supervised_vs_pseudolabeled_od}
    \vspace{-12px}
\end{table}

\begin{table}[h!]
\small
    \centering
    \rowcolors{2}{gray!15}{white}
    \begin{tabular}{ c | >{\centering}m{1.1cm} >{\centering}m{1.2cm} | >{\centering}m{0.4cm} >{\centering}m{0.4cm} >{\centering}m{0.4cm} c}
    Training & OD / Kir & OD / Kir & \multicolumn{4}{c}{Kirkland L1 AP} \\
    Method & \# label & \# pseudo & Veh  & $\Delta$ & Ped & $\Delta$ \\
    \midrule
    baseline & 800 / 0 & 0 / 0 & 41.8 & -- & 24.8 & -- \\
    supervised & 800 / 80 & 0 / 0 & 45.0 & {\footnotesize \textcolor{ForestGreen}{+3.2}} & 30.3 & {\footnotesize \textcolor{ForestGreen}{+5.5}} \\
    semi-super & 800 / 0 & \,\,\,\,\,\,0 / 560 & 44.5 & {\footnotesize \textcolor{ForestGreen}{+2.7}} & 28.4 & {\footnotesize \textcolor{ForestGreen}{+3.6}} \\
    semi-super & 800 / 0 & \,\,\,\,0 / 8k & \textbf{48.0} & {\footnotesize \textcolor{ForestGreen}{+6.2}} & 29.3 & {\footnotesize \textcolor{ForestGreen}{+4.5}} \\
    semi-super & 800 / 0 & 67k / 8k & \textbf{49.7} & {\footnotesize \textcolor{ForestGreen}{+7.9}} & 27.3 & {\footnotesize \textcolor{ForestGreen}{+2.5}} \\
    \end{tabular}
    \caption{\textbf{Pseudo-labeling out-of-domain data outperforms supervised training on new geographies.} 
    We report validation set L1 difficulty AP for Kirkland with the {\it same 1$\times$ width 1 frame} architecture, only varying the training method. $\Delta$ is the difference in AP with respect to the {\it Baseline} model. {\it Baseline} model is trained on Waymo Open Dataset (OD) but tested on a distinct geography (Kirkland). {\it Supervised} is trained on labeled data from OD and the distinct geography (Kirkland). {\it Semi-supervised} uses a 4$\times$ width 4 frame teacher model trained on OD labeled data to provide pseudo-labels, and trains on the joint labeled and pseudo-labeled data.}
    \label{tab:supervised_vs_pseudolabeled_kir}
    \vspace{-12px}
\end{table}

\subsection{Pushing unlabeled dataset size} \label{large_scale_unlabeled_data}

We return to our hypothesis that pseudo-labeling works best when the ratio of labeled data to unlabeled data is low. In practice, unlabeled self driving data is plentiful, so understanding how pseudo-labeling performs as the unlabeled dataset gets significantly larger is important. 

To scales the size of our unlabeled dataset, we were granted access to $>$100x more \textit{unlabeled data} from San Francisco (one of the three cities in the original Waymo Open Dataset) and Kirkland. This data contains $\sim$67,000 \textit{run segments} from San Francisco and $\sim$8,000 \textit{run segments} from Kirkland, as compared to the original 798 run segments from the original Waymo Open Dataset and 560 run segments from Kirkland. 

Empirically, we find that explicitly controlling the ratio of labeled to unlabeled data becomes important, as our unlabeled data otherwise overwhelms the labeled data. We train a 4$\times$ width, 4 frame teacher on the original Waymo Open Dataset, and use this to pseudo-label all $\sim$75,000 unlabeled run segments. We then train student models with a mix of all 798 labeled original Waymo Open Dataset run segments and a subset of these new pseudo-labeled run segments. While we did not exhaustively sweep the ratio of labeled-to-unlabeled data, in general we found a 1:5 ratio to work best (except for our Pedestrian model that used all $\sim$75,000 run segments, which worked best with a ratio of 1:1).

Our results show continued gains as we scale the amount of unlabeled data on both the Waymo Open Dataset (Table \ref{tab:supervised_vs_pseudolabeled_od}) and Kirkland (Table \ref{tab:supervised_vs_pseudolabeled_kir}).
 Vehicle models significantly improve, with a +5.8 AP improvement on the Waymo Open Dataset validation set, and a +7.9 AP improvement on the Kirkland validation set. For Pedestrians on the validation set, we see smaller gains of +1.5 AP on the original Waymo Open Dataset, and +4.5 on Kirkland, and more sensitivity to where the unlabeled data came from.  Our analysis shows that many scenes, especially in Kirkland, have very few or zero pedestrians\footnote{In the labeled validation sets, we found 70\% of scenes in the original Waymo Open Dataset had Pedestrians, with an average of 12.4 per scene, whereas in Kirkland only 22\% of scenes had Pedestrians, with an average of 0.57 per scene.}. We suspect that this introduces biases in the training process, and leave to future work to explore how to best choose \textit{which} pseudo-labeled frames to train on.
 
 We confirm these gains by evaluating on the test sets in Table \ref{tab:test_results}, where we achieve \textbf{state of the art accuracy} on both Vehicles and Pedestrians among all published single frame, LiDAR only, non-ensemble results available. We reiterate that we do not change the architecture, model hyperparameters, or training setup of the student; our only change is to add additional unlabeled data via pseudo-labeling.
 
\begin{table}[t!]
\small
    \centering
    \rowcolors{4}{gray!15}{white}
    \begin{tabular}{ c | c c | c  c}
    \multirow{2}{*}{Model} & \multicolumn{2}{c|}{Vehicle} & \multicolumn{2}{c}{Pedestrian} \\
    & L1 AP & L1 APH & L1 AP & L1 APH \\
    \midrule
    \multicolumn{5}{c}{\textbf{Waymo Open Dataset}} \\
    \midrule
    Second \cite{yan2018second} & 50.1 & 49.6 & -- & -- \\    
    StarNet \cite{ngiam2019starnet} & 63.5 & 63.0 & 67.8 & 60.1 \\
    PointPillars$^\dagger$ \cite{lang2019pointpillars} & 68.6 & 68.1 & 67.9 & 55.5 \\
    SA-SSD \cite{he2020structure} & 70.2 & 69.5 & 57.1 & 48.8 \\
    RCD \cite{bewley2020range} & 71.9 & 71.6 & -- & -- \\
    Ours$^\dagger$ & \textbf{74.0} & \textbf{73.6} & \textbf{69.8} & 57.9 \\
    \midrule
    \multicolumn{5}{c}{\textbf{Kirkland}} \\
    \midrule
    PointPillars$^\dagger$ \cite{lang2019pointpillars} & 49.3 & 48.8 & 37.5 & 29.7 \\
    Ours$^\dagger$ & \textbf{56.2} & \textbf{55.7} & 36.1 & 28.5 \\
    \end{tabular}
    \caption{\textbf{Test set results on the Waymo Open Dataset (top) and Kirkland Dataset (bottom).} We compare to other published single frame, LiDAR-only, non-ensemble methods. $\dagger$ indicates that both models were implemented, trained and evaluated by us, and are identical models in training setup and parameter count; the only difference is that our model was trained on $\sim$75k unlabeled run segments.}
    \label{tab:test_results}
    \vspace{-10px}
\end{table}

\subsection{Negative results}
Finally, we briefly touch on ideas that {\it did not work}, despite positive evidence in the literature for other tasks~\cite{xie2020self}. First, we tried two forms of soft labels, neither of which showed a gain. Second, we performed multiple iterations of training, which showed a small gain, but we deemed it too time-consuming to be worth it. 
Third, we explored whether there was an ambiguous range of classification scores between which we should assume the pseudo-object is neither labeled foreground or background, and anchors assigned to pseudo-label objects with these scores should receive no loss. We detail our experiments in Appendix \ref{apx:negative_results}.

\section{Conclusion}
Our work presents the first results of applying pseudo-label training to 3D object detection for self-driving car perception. We use a simple form of pseudo-labeling that requires no architecture innovation, yet when deployed in a semi-supervised learning paradigm, leads to substantial gains over supervised learning baselines on vehicle and pedestrian detection. Most interestingly, gains persist in the presence of domain shift and new environments where building new supervised label datasets has been a barrier to safe, wide deployment. Furthermore, we identify several prescriptions for maximizing pseudo-label-based training, including the construction of better teacher model architectures and leveraging data augmentation. To summarize our main results: 

\begin{compactitem}
    \item By distilling a large teacher model into a smaller student model and leveraging a large corpus of unlabeled data, we use a two year old architecture \cite{lang2019pointpillars} to achieve state-of-the-art results\footnote{When compared to other single-frame, LiDAR only, non-ensemble models.} of 74.0 / 69.8 L1 AP (+5.4 / +1.9 over supervised baseline) for Vehicles / Pedestrians, respectively, on the Waymo Open Dataset \textit{test set}.
    \item Using only 10\% of the labeled run segments, we show that Vehicle and Pedestrian student models can outperform equivalent supervised models trained with 3-10$\times$ as much labeled data, achieving a gain of 9.8 AP or larger for both classes and datasets.
    \item On the Kirkland Domain Adaptation Challenge, we show that pseudo-labeling produces more robust student models; our best model outperforms the equivalent supervised model by 7.9 / 4.5 L1 AP on the Kirkland validation set for Vehicles and Pedestrians, respectively.

\end{compactitem}

Overall, our work continues a long-standing theme of adapting unsupervised and semi-supervised learning techniques to problems in domain adaptation and the low label limit \cite{bousmalis2017unsupervised,shrivastava2017learning,berthelot2019mixmatch,berthelot2019remixmatch}. A majority of these methods have been tested on synthetic problems \cite{ganin2016domain,bousmalis2017unsupervised} or small academic datasets \cite{lecun2010mnist,krizhevsky2010convolutional}, and accordingly, such works leave open the question of how these methods may fare in the real-world. We suspect that domain adaptation in self-driving car perception may present a large-scale problem that may address such concerns and may help orient the semi-supervised learning field to a problem of critical importance for self-driving cars.

\iftoggle{cvprfinal}{
\section*{Acknowledgements}
We would like to thank Drago Anguelov, Shuyang Cheng, Ekin Dogus Cubuk, Barret Zoph, Rapha Gontijo Lopes, Wei Han, Zhaoqi Leng, Thang Luong, Charles Qi, Pei Sun, and Yin Zhou for helpful feedback on this work.
}{}

{\small
\bibliographystyle{ieee_fullname}
\bibliography{paper}

\begin{thebibliography}{10}\itemsep=-1pt

\bibitem{berthelot2019remixmatch}
David Berthelot, Nicholas Carlini, Ekin~D Cubuk, Alex Kurakin, Kihyuk Sohn, Han
  Zhang, and Colin Raffel.
\newblock Remixmatch: Semi-supervised learning with distribution alignment and
  augmentation anchoring.
\newblock {\em arXiv preprint arXiv:1911.09785}, 2019.

\bibitem{berthelot2019mixmatch}
David Berthelot, Nicholas Carlini, Ian Goodfellow, Nicolas Papernot, Avital
  Oliver, and Colin~A Raffel.
\newblock Mixmatch: A holistic approach to semi-supervised learning.
\newblock In {\em Advances in Neural Information Processing Systems}, pages
  5049--5059, 2019.

\bibitem{bewley2020range}
Alex Bewley, Pei Sun, Thomas Mensink, Dragomir Anguelov, and Cristian
  Sminchisescu.
\newblock Range conditioned dilated convolutions for scale invariant 3d object
  detection, 2020.

\bibitem{biggio2017wild}
Battista Biggio and Fabio Roli.
\newblock Wild patterns: Ten years after the rise of adversarial machine
  learning.
\newblock {\em Pattern Recognition}, 2018.

\bibitem{bousmalis2017unsupervised}
Konstantinos Bousmalis, Nathan Silberman, David Dohan, Dumitru Erhan, and Dilip
  Krishnan.
\newblock Unsupervised pixel-level domain adaptation with generative
  adversarial networks.
\newblock In {\em Proceedings of the IEEE conference on computer vision and
  pattern recognition}, pages 3722--3731, 2017.

\bibitem{caesar2020nuscenes}
Holger Caesar, Varun Bankiti, Alex~H Lang, Sourabh Vora, Venice~Erin Liong,
  Qiang Xu, Anush Krishnan, Yu Pan, Giancarlo Baldan, and Oscar Beijbom.
\newblock nuscenes: A multimodal dataset for autonomous driving.
\newblock In {\em Proceedings of the IEEE/CVF Conference on Computer Vision and
  Pattern Recognition}, pages 11621--11631, 2020.

\bibitem{Chen2020Leveraging}
Liang-Chieh Chen, Raphael~Gontijo Lopes, Bowen Cheng, Maxwell~D Collins, Ekin~D
  Cubuk, Barret Zoph, Hartwig Adam, and Jonathon Shlens.
\newblock Leveraging semi-supervised learning in video sequences for urban
  scene segmentation.
\newblock In {\em European Conference on Computer Vision (ECCV)}, 2020.

\bibitem{chen2020semi}
Liang-Chieh Chen, Raphael~Gontijo Lopes, Bowen Cheng, Maxwell~D Collins, Ekin~D
  Cubuk, Barret Zoph, Hartwig Adam, and Jonathon Shlens.
\newblock Semi-supervised learning in video sequences for urban scene
  segmentation.
\newblock {\em arXiv preprint arXiv:2005.10266}, 2020.

\bibitem{cheng2020improving}
Shuyang Cheng, Zhaoqi Leng, Ekin~Dogus Cubuk, Barret Zoph, Chunyan Bai, Jiquan
  Ngiam, Yang Song, Benjamin Caine, Vijay Vasudevan, Congcong Li, et~al.
\newblock Improving 3d object detection through progressive population based
  augmentation.
\newblock {\em arXiv preprint arXiv:2004.00831}, 2020.

\bibitem{dai2017scannet}
Angela Dai, Angel~X. Chang, Manolis Savva, Maciej Halber, Thomas Funkhouser,
  and Matthias Nießner.
\newblock Scannet: Richly-annotated 3d reconstructions of indoor scenes, 2017.

\bibitem{ding20201st}
Zhuangzhuang Ding, Yihan Hu, Runzhou Ge, Li Huang, Sijia Chen, Yu Wang, and Jie
  Liao.
\newblock 1st place solution for waymo open dataset challenge--3d detection and
  domain adaptation.
\newblock {\em arXiv preprint arXiv:2006.15505}, 2020.

\bibitem{ganin2016domain}
Yaroslav Ganin, Evgeniya Ustinova, Hana Ajakan, Pascal Germain, Hugo
  Larochelle, Fran{\c{c}}ois Laviolette, Mario Marchand, and Victor Lempitsky.
\newblock Domain-adversarial training of neural networks.
\newblock {\em The Journal of Machine Learning Research}, 17(1):2096--2030,
  2016.

\bibitem{ge2020afdet}
Runzhou Ge, Zhuangzhuang Ding, Yihan Hu, Yu Wang, Sijia Chen, Li Huang, and
  Yuan Li.
\newblock Afdet: Anchor free one stage 3d object detection.
\newblock {\em arXiv preprint arXiv:2006.12671}, 2020.

\bibitem{geiger2013vision}
Andreas Geiger, Philip Lenz, Christoph Stiller, and Raquel Urtasun.
\newblock Vision meets robotics: The kitti dataset.
\newblock {\em The International Journal of Robotics Research},
  32(11):1231--1237, 2013.

\bibitem{han2020streaming}
Wei Han, Zhengdong Zhang, Benjamin Caine, Brandon Yang, Christoph Sprunk, Ouais
  Alsharif, Jiquan Ngiam, Vijay Vasudevan, Jonathon Shlens, and Zhifeng Chen.
\newblock Streaming object detection for 3-d point clouds.
\newblock In {\em European Conference on Computer Vision (ECCV)}, 2020.

\bibitem{he2020structure}
Chenhang He, Hui Zeng, Jianqiang Huang, Xian-Sheng Hua, and Lei Zhang.
\newblock Structure aware single-stage 3d object detection from point cloud.
\newblock In {\em Proceedings of the IEEE/CVF Conference on Computer Vision and
  Pattern Recognition}, pages 11873--11882, 2020.

\bibitem{hendrycks2018benchmarking}
Dan Hendrycks and Thomas Dietterich.
\newblock Benchmarking neural network robustness to common corruptions and
  perturbations.
\newblock In {\em International Conference on Learning Representations (ICLR)},
  2019.

\bibitem{hinton2015distilling}
Geoffrey Hinton, Oriol Vinyals, and Jeff Dean.
\newblock Distilling the knowledge in a neural network, 2015.

\bibitem{houston2020one}
John Houston, Guido Zuidhof, Luca Bergamini, Yawei Ye, Ashesh Jain, Sammy
  Omari, Vladimir Iglovikov, and Peter Ondruska.
\newblock One thousand and one hours: Self-driving motion prediction dataset.
\newblock {\em arXiv preprint arXiv:2006.14480}, 2020.

\bibitem{kahn2020self}
Jacob Kahn, Ann Lee, and Awni Hannun.
\newblock Self-training for end-to-end speech recognition.
\newblock In {\em ICASSP 2020-2020 IEEE International Conference on Acoustics,
  Speech and Signal Processing (ICASSP)}, pages 7084--7088. IEEE, 2020.

\bibitem{kingma2014adam}
Diederik~P Kingma and Jimmy Ba.
\newblock Adam: A method for stochastic optimization.
\newblock {\em arXiv preprint arXiv:1412.6980}, 2014.

\bibitem{krizhevsky2010convolutional}
Alex Krizhevsky and Geoff Hinton.
\newblock Convolutional deep belief networks on cifar-10.
\newblock {\em Unpublished manuscript}, 40(7):1--9, 2010.

\bibitem{laine2016temporal}
Samuli Laine and Timo Aila.
\newblock Temporal ensembling for semi-supervised learning.
\newblock {\em arXiv preprint arXiv:1610.02242}, 2016.

\bibitem{lang2019pointpillars}
Alex~H Lang, Sourabh Vora, Holger Caesar, Lubing Zhou, Jiong Yang, and Oscar
  Beijbom.
\newblock Pointpillars: Fast encoders for object detection from point clouds.
\newblock In {\em Proceedings of the IEEE Conference on Computer Vision and
  Pattern Recognition}, pages 12697--12705, 2019.

\bibitem{lecun2010mnist}
Yann LeCun, Corinna Cortes, and CJ Burges.
\newblock Mnist handwritten digit database.
\newblock {\em ATT Labs [Online]. Available: http://yann.lecun.com/exdb/mnist},
  2, 2010.

\bibitem{lee2013pseudo}
Dong-Hyun Lee.
\newblock Pseudo-label: The simple and efficient semi-supervised learning
  method for deep neural networks.
\newblock In {\em Workshop on challenges in representation learning, ICML},
  volume~3, 2013.

\bibitem{li2010optimol}
Li-Jia Li and Li Fei-Fei.
\newblock Optimol: automatic online picture collection via incremental model
  learning.
\newblock {\em {IJCV}}, 2010.

\bibitem{li2020pointaugment}
Ruihui Li, Xianzhi Li, Pheng-Ann Heng, and Chi-Wing Fu.
\newblock Pointaugment: an auto-augmentation framework for point cloud
  classification.
\newblock In {\em Proceedings of the IEEE/CVF Conference on Computer Vision and
  Pattern Recognition}, pages 6378--6387, 2020.

\bibitem{luo2018fast}
Wenjie Luo, Bin Yang, and Raquel Urtasun.
\newblock Fast and furious: Real time end-to-end 3d detection, tracking and
  motion forecasting with a single convolutional net.
\newblock In {\em Proceedings of the IEEE Conference on Computer Vision and
  Pattern Recognition}, pages 3569--3577, 2018.

\bibitem{mclachlan1975iterative}
Geoffrey~J McLachlan.
\newblock Iterative reclassification procedure for constructing an
  asymptotically optimal rule of allocation in discriminant analysis.
\newblock {\em Journal of the American Statistical Association},
  70(350):365--369, 1975.

\bibitem{meng2020weakly}
Qinghao Meng, Wenguan Wang, Tianfei Zhou, Jianbing Shen, Luc Van~Gool, and
  Dengxin Dai.
\newblock Weakly supervised 3d object detection from lidar point cloud.
\newblock {\em arXiv preprint arXiv:2007.11901}, 2020.

\bibitem{ngiam2019starnet}
Jiquan Ngiam, Benjamin Caine, Wei Han, Brandon Yang, Yuning Chai, Pei Sun, Yin
  Zhou, Xi Yi, Ouais Alsharif, Patrick Nguyen, et~al.
\newblock Starnet: Targeted computation for object detection in point clouds.
\newblock {\em arXiv preprint arXiv:1908.11069}, 2019.

\bibitem{papandreou2015weakly}
George Papandreou, Liang-Chieh Chen, Kevin~P Murphy, and Alan~L Yuille.
\newblock Weakly-and semi-supervised learning of a deep convolutional network
  for semantic image segmentation.
\newblock In {\em {ICCV}}, 2015.

\bibitem{park2020improved}
Daniel~S Park, Yu Zhang, Ye Jia, Wei Han, Chung-Cheng Chiu, Bo Li, Yonghui Wu,
  and Quoc~V Le.
\newblock Improved noisy student training for automatic speech recognition.
\newblock {\em arXiv preprint arXiv:2005.09629}, 2020.

\bibitem{radosavovic2018data}
Ilija Radosavovic, Piotr Doll{\'a}r, Ross Girshick, Georgia Gkioxari, and
  Kaiming He.
\newblock Data distillation: Towards omni-supervised learning.
\newblock In {\em {CVPR}}, 2018.

\bibitem{recht2019imagenet}
Benjamin Recht, Rebecca Roelofs, Ludwig Schmidt, and Vaishaal Shankar.
\newblock Do imagenet classifiers generalize to imagenet?
\newblock In {\em International Conference on Machine Learning}, pages
  5389--5400, 2019.

\bibitem{rosenberg2005semi}
Chuck Rosenberg, Martial Hebert, and Henry Schneiderman.
\newblock Semi-supervised self-training of object detection models.
\newblock {\em WACV/MOTION}, 2005.

\bibitem{sajjadi2016regularization}
Mehdi Sajjadi, Mehran Javanmardi, and Tolga Tasdizen.
\newblock Regularization with stochastic transformations and perturbations for
  deep semi-supervised learning.
\newblock In {\em Advances in neural information processing systems}, pages
  1163--1171, 2016.

\bibitem{scudder1965probability}
H Scudder.
\newblock Probability of error of some adaptive pattern-recognition machines.
\newblock {\em IEEE Transactions on Information Theory}, 1965.

\bibitem{shrivastava2017learning}
Ashish Shrivastava, Tomas Pfister, Oncel Tuzel, Joshua Susskind, Wenda Wang,
  and Russell Webb.
\newblock Learning from simulated and unsupervised images through adversarial
  training.
\newblock In {\em Proceedings of the IEEE conference on computer vision and
  pattern recognition}, pages 2107--2116, 2017.

\bibitem{sohn2020fixmatch}
Kihyuk Sohn, David Berthelot, Chun-Liang Li, Zizhao Zhang, Nicholas Carlini,
  Ekin~D Cubuk, Alex Kurakin, Han Zhang, and Colin Raffel.
\newblock Fixmatch: Simplifying semi-supervised learning with consistency and
  confidence.
\newblock {\em arXiv preprint arXiv:2001.07685}, 2020.

\bibitem{sohn2020simple}
Kihyuk Sohn, Zizhao Zhang, Chun-Liang Li, Han Zhang, Chen-Yu Lee, and Tomas
  Pfister.
\newblock A simple semi-supervised learning framework for object detection.
\newblock {\em arXiv preprint arXiv:2005.04757}, 2020.

\bibitem{song2015sun}
Shuran Song, Samuel~P Lichtenberg, and Jianxiong Xiao.
\newblock Sun rgb-d: A rgb-d scene understanding benchmark suite.
\newblock In {\em Proceedings of the IEEE conference on computer vision and
  pattern recognition}, pages 567--576, 2015.

\bibitem{sun2020scalability}
Pei Sun, Henrik Kretzschmar, Xerxes Dotiwalla, Aurelien Chouard, Vijaysai
  Patnaik, Paul Tsui, James Guo, Yin Zhou, Yuning Chai, Benjamin Caine, et~al.
\newblock Scalability in perception for autonomous driving: Waymo open dataset.
\newblock In {\em Proceedings of the IEEE/CVF Conference on Computer Vision and
  Pattern Recognition}, pages 2446--2454, 2020.

\bibitem{szegedy2013intriguing}
Christian Szegedy, Wojciech Zaremba, Ilya Sutskever, Joan Bruna, Dumitru Erhan,
  Ian~J. Goodfellow, and Rob Fergus.
\newblock Intriguing properties of neural networks.
\newblock In {\em International Conference on Learning Representations (ICLR)},
  2013.

\bibitem{tang2019transferable}
Yew~Siang Tang and Gim~Hee Lee.
\newblock Transferable semi-supervised 3d object detection from rgb-d data.
\newblock In {\em Proceedings of the IEEE International Conference on Computer
  Vision}, pages 1931--1940, 2019.

\bibitem{tarvainen2018mean}
Antti Tarvainen and Harri Valpola.
\newblock Mean teachers are better role models: Weight-averaged consistency
  targets improve semi-supervised deep learning results, 2018.

\bibitem{thrun2006stanley}
Sebastian Thrun, Mike Montemerlo, Hendrik Dahlkamp, David Stavens, Andrei Aron,
  James Diebel, Philip Fong, John Gale, Morgan Halpenny, Gabriel Hoffmann,
  et~al.
\newblock Stanley: The robot that won the darpa grand challenge.
\newblock {\em Journal of field Robotics}, 23(9):661--692, 2006.

\bibitem{wang20203dioumatch}
He Wang, Yezhen Cong, Or Litany, Yue Gao, and Leonidas~J. Guibas.
\newblock 3dioumatch: Leveraging iou prediction for semi-supervised 3d object
  detection, 2020.

\bibitem{wang2020train}
Yan Wang, Xiangyu Chen, Yurong You, Li~Erran Li, Bharath Hariharan, Mark
  Campbell, Kilian~Q Weinberger, and Wei-Lun Chao.
\newblock Train in germany, test in the usa: Making 3d object detectors
  generalize.
\newblock In {\em Proceedings of the IEEE/CVF Conference on Computer Vision and
  Pattern Recognition}, pages 11713--11723, 2020.

\bibitem{wang2020multi}
Yue Wang, Alireza Fathi, Jiajun Wu, Thomas Funkhouser, and Justin Solomon.
\newblock Multi-frame to single-frame: Knowledge distillation for 3d object
  detection.
\newblock {\em arXiv preprint arXiv:2009.11859}, 2020.

\bibitem{xie2020self}
Qizhe Xie, Minh-Thang Luong, Eduard Hovy, and Quoc~V Le.
\newblock Self-training with noisy student improves imagenet classification.
\newblock In {\em Conference on Computer Vision and Pattern Recognition
  (CVPR)}, 2020.

\bibitem{billion_large_scale}
I.~Zeki Yalniz, Herv{'e} J{'e}gou, Kan Chen, Manohar Paluri, and Dhruv Mahajan.
\newblock Billion-scale semi-supervised learning for image classification.
\newblock {\em arXiv 1905.00546}, 2019.

\bibitem{yan2018second}
Yan Yan, Yuxing Mao, and Bo Li.
\newblock Second: Sparsely embedded convolutional detection.
\newblock {\em Sensors}, 18(10):3337, 2018.

\bibitem{yang2021auto4d}
Bin Yang, Min Bai, Ming Liang, Wenyuan Zeng, and Raquel Urtasun.
\newblock Auto4d: Learning to label 4d objects from sequential point clouds,
  2021.

\bibitem{yang2018hdnet}
Bin Yang, Ming Liang, and Raquel Urtasun.
\newblock Hdnet: Exploiting {HD} maps for 3d object detection.
\newblock In {\em Conference on Robot Learning}, pages 146--155, 2018.

\bibitem{yang2018pixor}
Bin Yang, Wenjie Luo, and Raquel Urtasun.
\newblock Pixor: Real-time 3d object detection from point clouds.
\newblock In {\em Proceedings of the IEEE Conference on Computer Vision and
  Pattern Recognition}, pages 7652--7660, 2018.

\bibitem{zhao2020sess}
Na Zhao, Tat-Seng Chua, and Gim~Hee Lee.
\newblock Sess: Self-ensembling semi-supervised 3d object detection, 2020.

\bibitem{zhou2020end}
Yin Zhou, Pei Sun, Yu Zhang, Dragomir Anguelov, Jiyang Gao, Tom Ouyang, James
  Guo, Jiquan Ngiam, and Vijay Vasudevan.
\newblock End-to-end multi-view fusion for 3d object detection in lidar point
  clouds.
\newblock In {\em Conference on Robot Learning}, pages 923--932, 2020.

\bibitem{zhou2018voxelnet}
Yin Zhou and Oncel Tuzel.
\newblock Voxelnet: End-to-end learning for point cloud based 3d object
  detection.
\newblock In {\em Proceedings of the IEEE Conference on Computer Vision and
  Pattern Recognition}, pages 4490--4499, 2018.

\bibitem{zoph2020rethinking}
Barret Zoph, Golnaz Ghiasi, Tsung-Yi Lin, Yin Cui, Hanxiao Liu, Ekin~D Cubuk,
  and Quoc~V Le.
\newblock Rethinking pre-training and self-training.
\newblock {\em arXiv preprint arXiv:2006.06882}, 2020.

\end{thebibliography}
}

\clearpage
\section{Appendix}
\subsection{Model and training details}
\subsubsection{PointPillars architecture}
\label{sec:pillar_details}

We use the "Pedestrian" version of the PointPillars architecture for both the Vehicle and the Pedestrian classes, which uses a stride of 1 for the first convolutional block (instead of 2). 
This results in the output resolution matching the input resolution, which we found important for maintaining accuracy scaling PointPillars to larger scenes. We adopt a resolution of 512 pixels, spanning [-76.8m, 76.8m] in both X and Y, and a Z range of [-3m, 3m] giving us a pixel size of 0.33m, which is similar to what is used by \cite{zhou2020end} on the Waymo Open Dataset. Additionally, on all models, we replace hard voxelization, which samples a fixed number of points per voxel, with a dynamic voxelization \cite{zhou2020end}, which allows the model to use all the points in the point cloud, and makes it efficiently able to handle larger point clouds. Adding dynamic voxelization has negligible effect on accuracy. 

\subsubsection{Training details}
\label{sec:training_details}

We use the Adam optimizer with an initial learning rate of 3.2e-3. We train for a total of 75 epochs with a batch size of 64. An exponential decay schedule of the learning rate starts at epoch 5. For models trained with 10\% of the original Waymo Open Dataset labeled run segments, we double the training time, so that the total epoch is 150, and the exponential decay starts at epoch 10. Lastly, on the large scale experiments in section \ref{large_scale_unlabeled_data}, we train for 15 total epochs, with our exponential decay starting at epoch 2. We apply an exponential moving average (EMA) decay of 0.99 on all variables and use L2 regularization with scaling constant 1e-4. 

Our anchor box prior corresponds to the mean box dimensions for each class and is [4.725, 2.079, 1.768] and [0.901, 0.857, 1.712] for Vehicles and Pedestrians respectively. Our anchors have two rotations of [0, $\pi/2$], and are placed in the middle of each voxel. In order to compute the loss function during training, we assign an anchor to a ground truth box if its IoU is greater than 0.6 for Vehicles, and 0.5 for Pedestrians, and to background if the IoU is below 0.45 for vehicles, and 0.35 for pedestrians. Boxes with IoU between these values have a loss weight of 0, and we use force matching to make sure every ground truth is assigned at least one box.

Unless specified otherwise, all students and teachers are trained with two data augmentations: RandomWorldRotationAboutZAxis, and RandomFlipY. For RandomWorldRotationAboutZAxis we choose a random rotation of up to $\pi/4$ to apply to the world around the Z axis. For RandomFlipY, we flip the Y coordinate, which can be thought of as mirroring the scene over the X axis, with probability of 0.25. 

\subsection{Is data augmentation necessary?}
\label{apx:data_augmentation_necessary}
\begin{table}[h!]
    \centering
    \rowcolors{2}{gray!15}{white}
    \begin{tabular}{c c |c c c}
      \multicolumn{2}{c |}{\textbf{Aug.?}} & \multicolumn{3}{c}{\textbf{OD / Kirkland AP}} \\
    \midrule
    Teach. & Stud. & Teacher & Student & $\Delta$ \\
    \midrule
    No & No  &  56.3 / 33.6 & 59.2 / 38.0 & \textcolor{ForestGreen}{+2.9} / \textcolor{ForestGreen}{+4.4} \\
    No & Yes &  56.3 / 33.6 & 62.2 / 40.9 & \textcolor{ForestGreen}{+5.9} / \textcolor{ForestGreen}{+7.4} \\
    Yes & No &  63.0 / 41.8 & 59.5 / 39.5 & \textcolor{red}{-3.5} / \textcolor{red}{-2.3} \\
    Yes & Yes & 63.0 / 41.8 & 64.2 / 44.0 & \textcolor{ForestGreen}{+1.2} / \textcolor{ForestGreen}{+2.2} \\
    \end{tabular}
    \caption{\textbf{Data augmentation is not necessary, but beneficial.} While data augmentation is not necessary, the best student is achieved when both the student and teacher receive the same advantages. Results are on 1x width, 1 frame vehicle models where both the teacher and student saw 100\% of the original Waymo Open Dataset labeled run segments.}
    \label{tab:teacher_student_augmentations}
\end{table}

\subsection{Kirkland results}
\label{apx:kirkland_teacher_configs}

We provide all the corresponding Kirkland validation set figures on Vehicles for Section \ref{better_teachers_better_students}. We show that all of our results shown for the Waymo Open Dataset still hold when we evaluate on Kirkland.

\begin{figure}[t!]
  \newcommand{\ppwm}{0.47}
  \centering
  \begin{subfigure}{\ppwm\textwidth}
  \centering
    \includegraphics[width=0.8\linewidth]{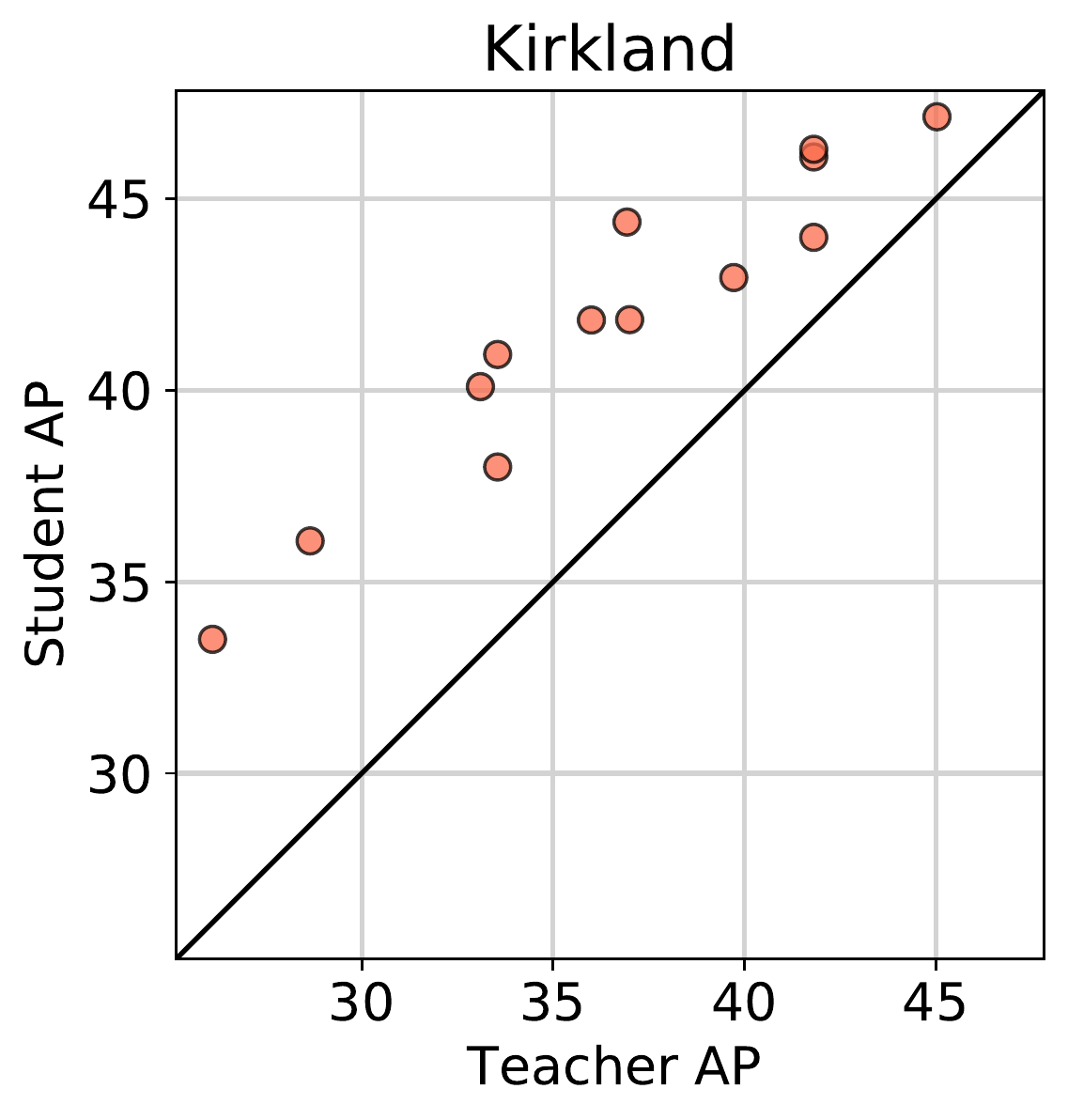}
  \end{subfigure}
  \caption{
    \textbf{Kirkland evaluation: Better teachers lead to better students.}
    If the student model has an equivalent architecture or training setup compared to the teacher, teachers with a higher AP produce students with a higher AP. All numbers are Vehicle models reporting Level 1 AP on the Kirkland \textit{validation split}.
    }
  \label{fig:student_vs_teacher_kirkland}
\end{figure}

\textbf{Better teachers lead to better students.} Similar to Figure \ref{fig:student_vs_teacher}, Figure \ref{fig:student_vs_teacher_kirkland} shows that improving the teacher accuracy leads to a corresponding increase in student accuracy.

\begin{figure}[t!]
  \newcommand{\ppwm}{0.48}
  \centering
    \includegraphics[width=\linewidth]{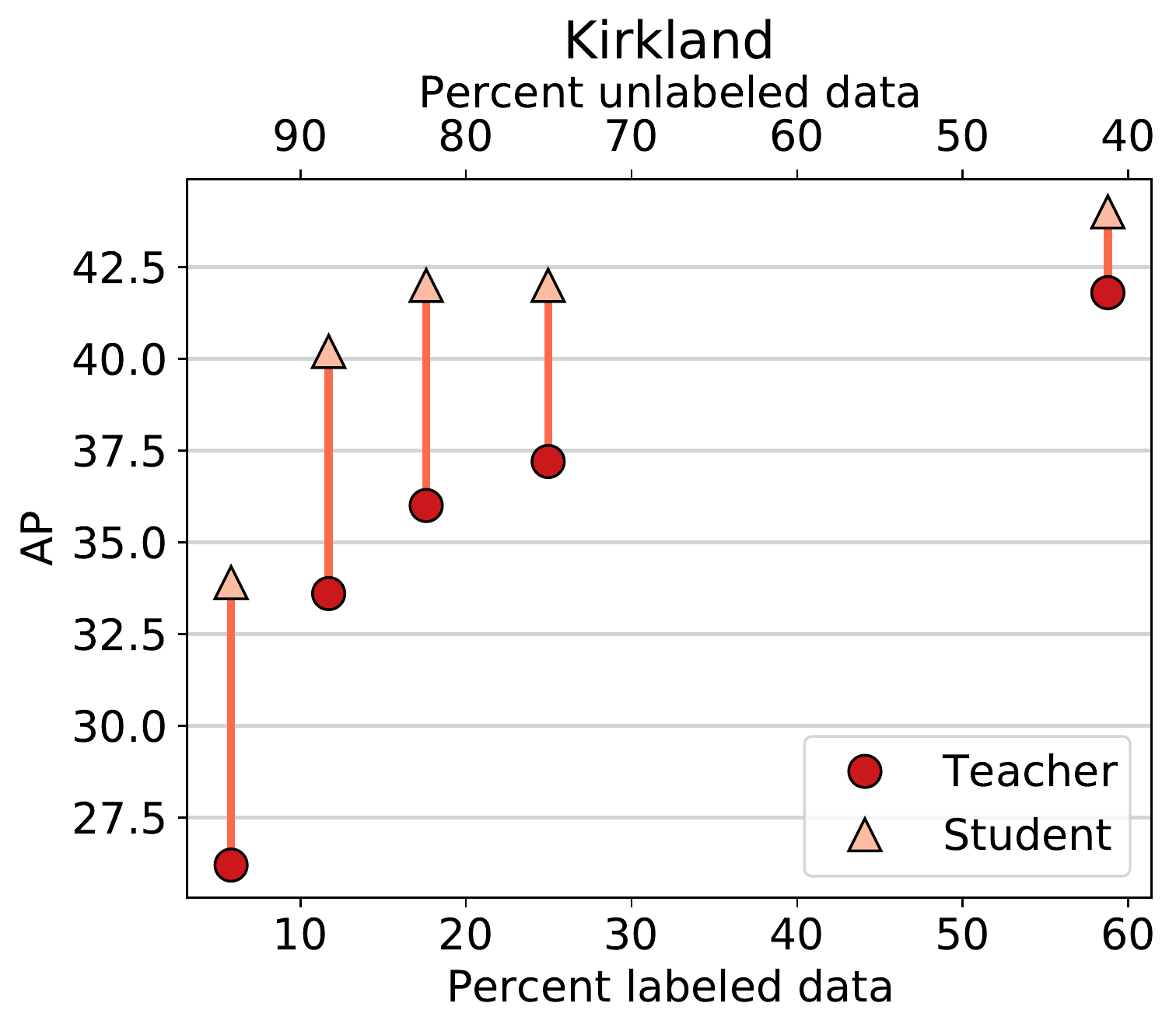}
  \caption{
    \textbf{Kirkland evaluation: Pseudo-label training is most effective when the ratio of labeled to unlabeled data is small.} Teacher and student L1 AP on the Waymo Open Dataset \textit{validation set} for the Vehicle class versus percent labeled data.
    }
  \label{fig:teacher_overall_percent_labeled_data_kirkland}
\end{figure}

\textbf{Amount of labeled data.} Again mirroring our results in Figure \ref{fig:teacher_overall_percent_labeled_data}, Figure \ref{fig:teacher_overall_percent_labeled_data_kirkland} shows increasing the amount of labeled data increases both the teacher and student performance. We also see similar (though less severe) diminishing returns as the ratio of labeled to unlabeled data gets larger. Both teachers and students are 1x width, 1 frame in these experiments.

\textbf{Data augmentations.}
We also show the equivalent of Table \ref{tab:teacher_augmentations_kirkland} when evaluating on Kirkland:
\begin{table}[t!]
    \centering
    \rowcolors{2}{gray!15}{white}
    \begin{tabular}{ c |c  c c }
      & \multicolumn{3}{c}{\textbf{Kirkland L1 AP}} \\ %
    \midrule
    Teacher Augmentation & Teacher & Student & $\Delta$ \\
    \midrule
    None    &  33.6  &  40.9  &  \textcolor{ForestGreen}{+7.3} \\
    \rotatez & 39.7 &  43.0  &  \textcolor{ForestGreen}{+3.3} \\
    \flipy   &  36.9 &  44.4 &  \textcolor{ForestGreen}{+7.5} \\
    \rotatez + \flipy & 41.8 & 44.0  & \textcolor{ForestGreen}{+2.2} \\
    \bottomrule
    \end{tabular}
    \caption{\textbf{Increasing the strength of teacher augmentations leads to additive gains in student performance.} We increasing the strength of teacher augmentations for a 1$\times$ width teacher model trained on 100\% of the Waymo Open Dataset. The student model is a fixed 1$\times$ width model trained with both \rotatez and \flipy augmentations. We report L1 \textit{validation set} Vehicle AP on the Kirkland dataset. $\Delta$ is the difference in AP between the student model and the teacher model.}
    \label{tab:teacher_augmentations_kirkland}
    \vspace{-0.2cm}
\end{table}

\begin{figure}[t!]
  \newcommand{\ppwm}{0.48}
  \centering
  \includegraphics[width=\linewidth]{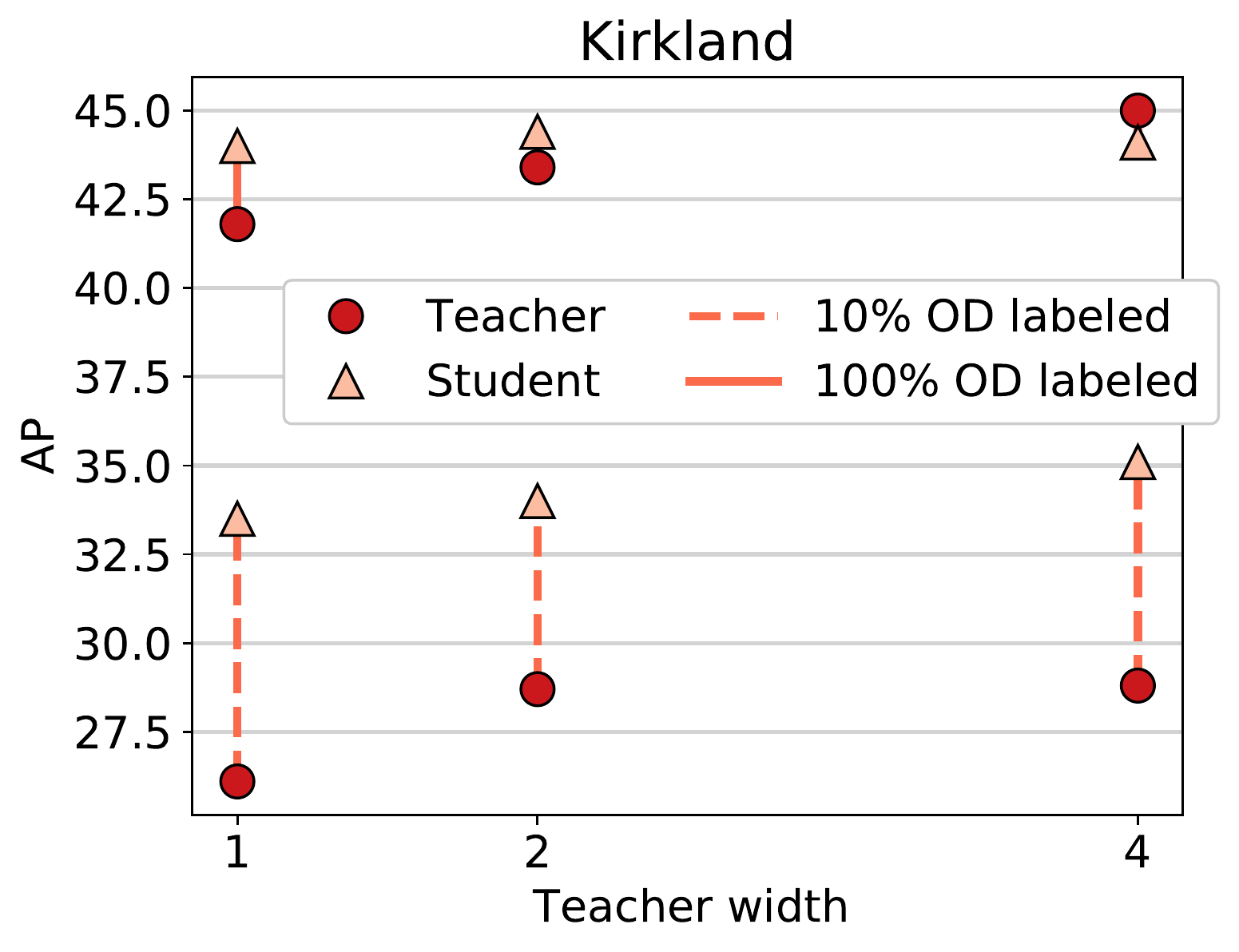}
  \caption{
    \textbf{Kirkland evaluation: Increasing teacher width leads to better students.}
    We make teachers wider while fixing the student width at 1x and report L1 AP for Vehicle models on the Kirkland \textit{validation set}. The teacher is trained on labeled data from either 10\% or 100\% of the original Waymo Open Dataset (top and bottom points, respectively).
    The student is trained on the labeled data seen by the teacher plus all unlabeled data from the Waymo Open Dataset and its Kirkland dataset. When the ratio of labeled to unlabeled data is small, the student accuracy improves as the teacher gets wider, however this effect disappears when the amount of pseudo-labeled data is small. We further investigate this by adding more unlabeled data in Section \ref{large_scale_unlabeled_data}.
    }
  \label{fig:teacher_width_kirkland}
\end{figure}

\textbf{Teacher width.}
Figure \ref{fig:teacher_width_kirkland} shows the effect of increasing teacher width for teachers trained on either 10\% or 100\% of the Waymo Open Dataset. We see the same result as in Figure \ref{fig:teacher_width}, where when we hold the student configuration fixed at 1x width, 1 frame, increasing the teacher width leads to an increase in student accuracy in the low data regime (10\% of original Waymo Open Dataset run segments). We see for Kirkland that similar to the original Waymo Open Dataset, when the ratio of labeled to unlabeled data is large (100\% of original Waymo Open Dataset run segments), this effect disappears. 

\subsection{Pushing labeled data efficiency}
\label{apx:pushing_labeled_data_efficiency}
Here we provide additional results where we push the accuracy of our student models on a limited amount of data organized as run segments. We replicated the experiment shown in Figure \ref{fig:finale_10pct} using 20\% of the original Waymo Open Dataset run segments (so $\sim$11.4\% of the overall run segments are labeled), and show similar gains. Additionally, we provide raw numerical values for all data points from these 8 plots in in Table \ref{tab:vehicle_finale} and Table \ref{tab:ped_finale}, to allow others to compare against us.

\begin{figure*}[t]
  \centering
  \includegraphics[width=\linewidth]{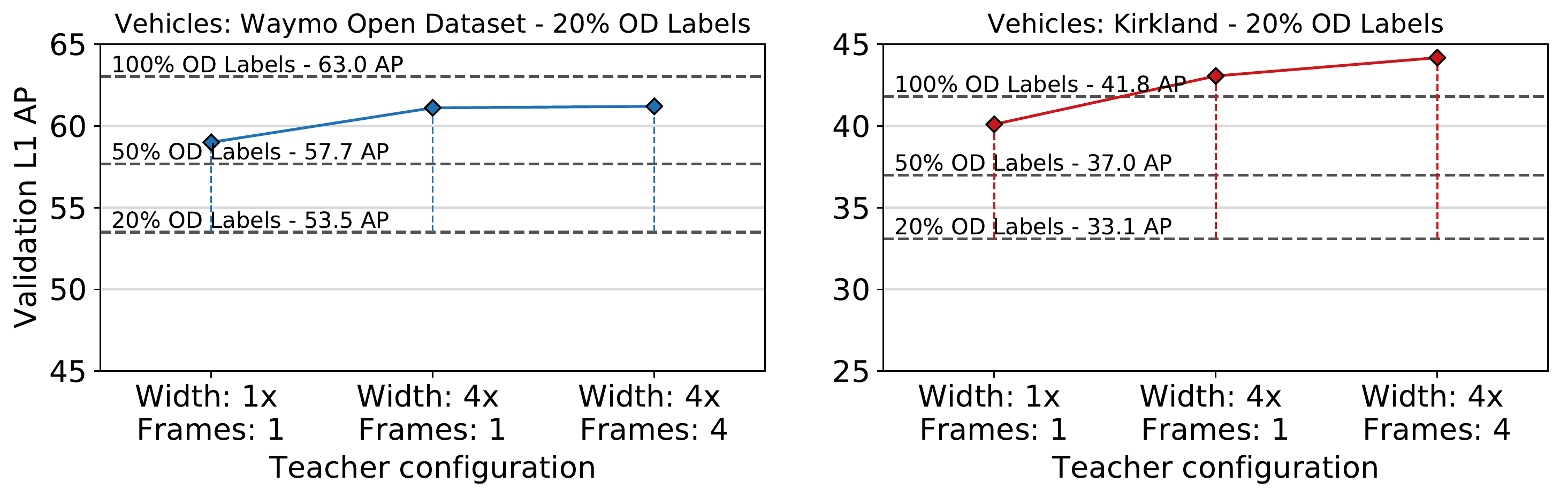}
  \includegraphics[width=\linewidth]{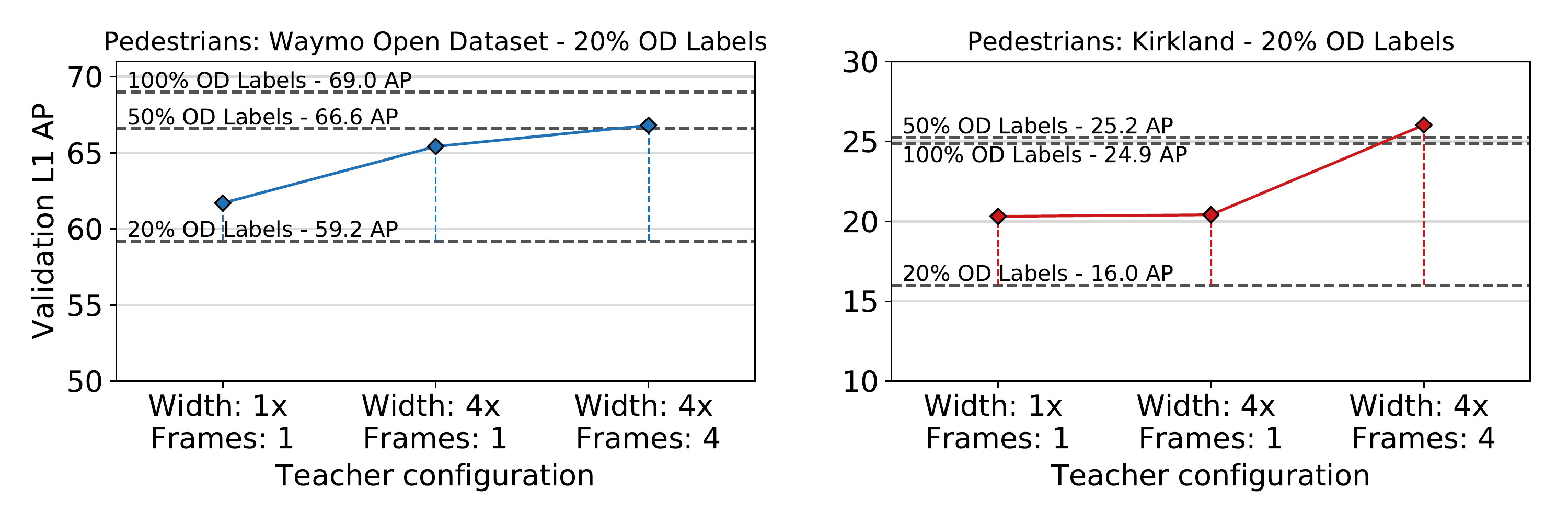}
  \vspace{-0.7cm}
  \caption{\textbf{Increasingly large teachers distill into small, accurate students.} Increasing the width or number of frames for the teacher impacts performance on a fixed size student (1$\times$ width, 1 frame) in the low label  (20\% of run segments) regime for vehicles (top) and pedestrians (bottom). Training a large, expensive teacher model, and distilling the teacher into a small, efficient student is efficient tactic. Results presented for Waymo Open Dataset (left) and Kirkland (right).}
  \label{fig:finale_20pct}
  \end{figure*}

\begin{table*}[]
    \centering
    \rowcolors{2}{gray!15}{white}
    \begin{tabular}{c c c |c c c c }
      \multicolumn{3}{c |}{\textbf{Model Details}} & \multicolumn{4}{c}{\textbf{OD / Kirkland L1 AP}} \\
    \midrule
    Teacher & Student & \% OD Labels & Teacher & Baseline & Student & $\Delta$ Baseline \\
    \midrule
    1x Width, 1 Frame & 1x Width, 1 Frame  &  10 & 49.1 / 26.1 & 49.1 / 26.1 & 54.6 / 33.5 & +5.5 / +7.4 \\
    4x Width, 1 Frame & 1x Width, 1 Frame &  10 & 52.2 / 28.7 & 49.1 / 26.1 & 57.7 / 35.3 & +8.6 / +9.2 \\
    4x Width, 4 Frame & 1x Width, 1 Frame &  10 & 54.1 / 30.4 & 49.1 / 26.1 & 58.9 / 37.2 & +9.8 / +11.1 \\
    1x Width, 1 Frame & 1x Width, 1 Frame & 20 & 53.5 / 33.1 & 53.5 / 33.1 & 59.0 / 40.1 & +5.5 / +7.0 \\
    4x Width, 1 Frame & 1x Width, 1 Frame & 20 & 58.6 / 38.4 & 53.5 / 33.1 & 61.1 / \textbf{43.1} & +7.6 / +10.0 \\
    4x Width, 4 Frame & 1x Width, 1 Frame & 20 & 60.0 / 39.8 & 53.5 / 33.1 & 61.2 / \textbf{44.2} & +7.7 / +11.1 \\
    \midrule
    1x Width, 1 Frame & & 30 & 56.4 / 36.0 \\
    1x Width, 1 Frame & & 50 & 57.7 / 37.0 \\
    1x Width, 1 Frame & & 100 & 63.0 / 41.8 \\
    \end{tabular}
    \caption{Vehicle results for single frame, normal width student models trained with increasingly complex (wider, multi-frame) teacher models. We show how it is advantageous to distill a complex, off-board model into a simple onboard model using pseudo-labeling. All numbers are on the corresponding \textit{Validation} set, and are Level 1 difficulty mean average precision (AP).}
    \label{tab:vehicle_finale}
\end{table*}

\begin{table*}[]
    \centering
    \rowcolors{2}{gray!15}{white}
    \begin{tabular}{c c c |c c c c }
      \multicolumn{3}{c |}{\textbf{Model Details}} & \multicolumn{4}{c}{\textbf{OD / Kirkland L1 AP}} \\
    \midrule
    Teacher & Student & \% OD Labels & Teacher & Baseline & Student & $\Delta$ Baseline \\
    \midrule
    1x Width, 1 Frame & 1x Width, 1 Frame  &  10 & 53.4 / 14.5 & 53.4 / 14.5 & 58.8 / 19.1 & +5.4 / +4.6 \\
    4x Width, 1 Frame & 1x Width, 1 Frame &  10 & 59.2 / 21.5 & 53.4 / 14.5  & 61.4 / 20.9 & +8.0 / +6.4 \\
    4x Width, 4 Frame & 1x Width, 1 Frame &  10 & 64.0 / 27.6 & 53.4 / 14.5 & 64.6 / \textbf{27.1} & +11.2 / +12.6 \\
    1x Width, 1 Frame & 1x Width, 1 Frame & 20 & 59.2 / 16.0 & 59.2 / 16.0 & 61.7 / 20.3 & +2.5 / +4.3 \\
    4x Width, 1 Frame & 1x Width, 1 Frame & 20 & 64.4 / 22.5 & 59.2 / 16.0 & 65.4 / 20.4 & +6.2 / +4.4 \\
    4x Width, 4 Frame & 1x Width, 1 Frame & 20 & 68.8 / 30.8 & 59.2 / 16.0 & 66.8 / \textbf{26.0} & +7.6 / +10.0 \\
    \midrule
    1x Width, 1 Frame & & 30 & 62.3 / 23.3 \\
    1x Width, 1 Frame & & 50 & 66.6 / 25.3 \\
    1x Width, 1 Frame & & 100 & 69.0 / 24.8\\
    \end{tabular}
    \caption{Pedestrian results for single frame, normal width student models trained with increasingly complex (wider, multi-frame) teacher models. We show how it is advantageous to distill a complex, off-board model into a simple onboard model using pseudo-labeling. All numbers are on the corresponding \textit{Validation} set, and are Level 1 difficulty mean average precision (AP).}
    \label{tab:ped_finale}
\end{table*}

\subsection{Different Teacher and Student Architectures}
In the main text we show that the teacher and student architecture can be different configurations, and in fact using a larger teacher is an effective way to generate significantly stronger, small student models. One remaining question is whether the teacher and student architectures need to be from the same architecture family, or even similar in their data representation. To test this, we design a very simple experiment where we take our best 10\% original Waymo OD run segment PointPillars teacher model (the exact model used in Figures \ref{fig:intro-figure} \& \ref{fig:finale_10pct}), and use it to pseudo label the remaining Waymo Open Dataset. We then train a StarNet \cite{ngiam2019starnet} student model on the union of the 10\% labeled run segments, and the remaining data pseudo labeled by PointPillars. We chose StarNet because it's a purely point-cloud based, convolution free object detection system, which differs significantly from PointPillars convolution-based architecture. Results are summarized in Table \ref{tab:starnet_results}, which shows strong gains in StarNet accuracy when using a PointPillars teacher.

\begin{table}[t!]
\small
    \centering
    \rowcolors{4}{gray!15}{white}
    \begin{tabular}{ c | c c | c  c}
    \multirow{2}{*}{Model} & \multicolumn{2}{c|}{Vehicle} & \multicolumn{2}{c}{Pedestrian} \\
    & L1 AP & $\Delta$ & L1 AP & $\Delta$ \\
    \midrule
    \multicolumn{5}{c}{\textbf{Waymo Open Dataset}} \\
    \midrule
    StarNet 10\% Baseline & 47.7 & - & 61.2 & - \\
    StarNet Student & 55.6 & \textcolor{ForestGreen}{+7.9} & 66.5 & \textcolor{ForestGreen}{+4.3} \\
    \midrule
    \multicolumn{5}{c}{\textbf{Kirkland}} \\
    \midrule
    StarNet 10\% Baseline & 26.3 & - & 6.7 & - \\
    StarNet Student & 35.2 & \textcolor{ForestGreen}{+8.9} & 22.2 & \textcolor{ForestGreen}{+15.5} \\
    \end{tabular}
    \caption{\textbf{Pseudo Labeling  is effective across very different architectures: } We distill a 4$\times$ width, 4 frame PointPillars teacher model into a single frame StarNet model and see large gains in StarNet performance, despite it being an extremely different architecture.}
    \label{tab:starnet_results}
    \vspace{-10px}
\end{table}

\subsection{Negative result details}
\label{apx:negative_results}
In this section, we provide some more details about our negative results.

\textbf{Soft-labels:} We explored two forms of soft labels, one of which was to use the post-sigmoid score bounded between [0, 1] as the target, the second was to use the logit itself. In object detection, because the outputs are passed through Non-Maximum Suppression (NMS), we only have scores and logits for \textit{foreground} locations, therefore background anchors all were assigned a score or logit of 1. We found both techniques resulted in \textit{slightly} worse performance than simply using hard labels. 

\textbf{Multiple iterations:} We tried multiple iteration training, where we used the best student checkpoint to re-pseudo-label the unlabeled data, and use that updated pseudo-labeled data to train a new student. While our trend thus far has shown better teachers lead to better students, its challenging to combat the overfitting that will naturally occur. Its our understanding that this is one of the main reasons one wants to heavily noise the student \cite{xie2020self}, but we found it difficult to find a noise level (via augmentations) that did not hamper model performance enough such that the second iteration was not worse. With default settings using the same augmentations for both the teacher, the first student, and the second student, we found a small gain in performance using 10\% of the original Waymo Open Dataset run segments of $\sim$0.2 AP on the original validation set, and $\sim$1.0 AP on the Kirkland validation set. Because of the small gains compared to the first iteration, and the time-consuming nature of performing these experiments, we left further exploration to future work.

\textbf{Score thresholds:} We wondered whether there may be some classification score range for pseudo-labels for which the class is ambiguous and we should assign no loss. We allowed anchors to be assigned a loss of zero if these anchors matched (via normal IoU matching) pseudo-label objects with scores between some [lower, upper] range. We then swept these two values, and found the most effective results were when both values were [0.5, 0.5], indicating this setting should be turned off. That said, we think the idea of limiting the noise induced by bad pseudo-labels merits future investigation.

\end{document}